\documentclass[nohyperref]{article}

\usepackage{microtype}
\usepackage{graphicx}
\usepackage{subfigure}
\usepackage{booktabs} 

\usepackage{hyperref}

\usepackage[accepted]{icml2022}

\usepackage{amsmath}
\usepackage{amssymb}
\usepackage{mathtools}
\usepackage{amsthm}

\usepackage[capitalize,noabbrev]{cleveref}

\theoremstyle{plain}

\theoremstyle{definition}

\theoremstyle{remark}

\usepackage[textsize=tiny]{todonotes}

\icmltitlerunning{Optimally Controllable Perceptual Lossy Compression}

\begin{document}

\twocolumn[
\icmltitle{Optimally Controllable Perceptual Lossy Compression}




\begin{icmlauthorlist}
\icmlauthor{Zeyu Yan}{edu}
\icmlauthor{Fei Wen}{edu}
\icmlauthor{Peilin Liu}{edu}
\end{icmlauthorlist}

\icmlaffiliation{edu}{Brain-inspired Application Technology Center (BATC), School of Electronic Information and Electrical Engineering, Shanghai Jiao Tong University, Shanghai, China}

\icmlcorrespondingauthor{Fei Wen}{wenfei@sjtu.edu.cn}

\icmlkeywords{Machine Learning, ICML}

\vskip 0.3in
]



\printAffiliationsAndNotice{}  

\begin{abstract}
Recent studies in lossy compression show that 
distortion and perceptual quality are at odds with each other, 
which put forward the tradeoff between distortion and perception (D-P). 
Intuitively, to attain different perceptual quality, 
different decoders have to be trained. 
In this paper, we present a nontrivial finding that 
only two decoders are sufficient for optimally achieving arbitrary 
(an infinite number of different) D-P tradeoff. 
We prove that arbitrary points of the D-P tradeoff bound can be achieved 
by a simple linear interpolation between the outputs of a minimum MSE decoder 
and a specifically constructed perfect perceptual decoder. 
Meanwhile, the perceptual quality 
(in terms of the squared Wasserstein-2 distance metric) 
can be quantitatively controlled by the interpolation factor. 
Furthermore, to construct a perfect perceptual decoder, 
we propose two theoretically optimal training frameworks. 
The new frameworks are different from the distortion-plus-adversarial loss 
based heuristic framework widely used in existing methods, 
which are not only theoretically optimal but also can yield 
state-of-the-art performance in practical perceptual decoding. 
Finally, we validate our theoretical finding and demonstrate 
the superiority of our frameworks via experiments. Code is available at: \textit{https://github.com/ZeyuYan/Controllable-Perceptual-Compression}
\end{abstract}

\section{Introduction}
\label{submission}

Lossy compression is an essential technique for digital data storage and transport. 
For decades, the goal of lossy compression 
is to achieve the lowest possible distortion at a given bit rate, 
which is bounded by Shannon's rate-distortion function \cite{1959Coding,ThomasCover2nd}.
However, recent studies show that typical distortion measures, 
e.g., MSE, PSNR and SSIM/MS-SSIM \cite{2003Multiscale,2004ImageQuality}, 
are not fully consistent with human’s subjective judgement on perceptual quality 
\cite{2016Perceptual, 2018Unreasonable, 2018PDtradeoff}.
The work \cite{2018PDtradeoff} lays a first theoretical foundation for understanding
the distortion-perception (D-P) tradeoff, revealing that distortion and perception 
quality are at odds with each other.
It accords well with empirical results that improving perceptual quality 
by adversarial learning and/or deep features based perceptual loss would lead to
an increase of the distortion \cite{2017DeepGenerative, 2018DeepGenerative, 2018Generative, 2019Extreme,2020Fidelity,2020HighFidelity, 2021highperceptualdenoise, 2021removepixelnoise}.

For lossy compression, the “classic" rate-distortion theory has been recently expanded
in \cite{2019Rethinking} to take the perceptual quality into consideration, 
resulting in a rate-distortion-perception three-way tradeoff.
Using a divergence from the source distribution
to measure the perceptual quality, the monotonicity and convexity of the 
rate-distortion-perception function have been established.
More recently, \cite{2021YanZy} shows that an optimal encoder for the “classic” rate-distortion 
problem is also optimal for the perceptual compression problem with perception constraint. 
Hence, a minimum MSE (MMSE) encoder is universal for perceptual lossy compression. In light of this, 
to obtain different perceptual quality, a natural way is to use a fixed MMSE encoder whilst 
train different decoders to fulfill different D-P tradeoff \cite{2021UniversalDP}. 

In this paper, we present a nontrivial theoretical finding that it is unnecessary to train 
different decoders for different perceptual quality, and two specifically constructed decoders 
are enough for achieving arbitrary D-P tradeoff. Furthermore, we propose two optimal training 
frameworks for perfect perceptual decoding to enable the realization of this kind of convenient D-P tradeoff.

In summary, our contributions are as follows:

\vspace{-1em}
\begin{itemize}
	\item[$\bullet$]
A theoretical finding for optimal D-P tradeoff in lossy compression. 
We prove that, with distortion measured by squared error and perceptual 
quality measured by squared Wasserstein-2 distance, arbitrary points of 
the D-P tradeoff bound can be achieved by a simple linear interpolation 
between an MMSE decoder and a specifically constructed perceptual decoder 
with perfect perception constraint. Besides, in theory the perceptual quality 
(measured in terms of squared Wasserstein-2 distance) can be quantitatively 
controlled by the interpolation factor. This finding reveals that two 
specifically constructed and paired decoders are sufficient for optimally 
achieving arbitrary (an infinite number of different) D-P tradeoff.
\item[$\bullet$]\vspace{-0.3em}
Two optimal training frameworks for perfect perceptual decoding, which 
enables the realization of interpolation based optimal D-P tradeoff. 
One of them trains a perceptual decoder to decode directly from compressed 
representation with the aid of an MMSE decoder. The other uses a combination 
structure consists of an MMSE encoder-decoder pair followed by a post-processing 
perceptual decoder. While the former has a more compact structure, the latter is 
more flexible that applies to existing lossy compression systems optimized 
only in terms of distortion measures. 
Particularly, while existing frameworks depend heavily on 
the VGG loss and hence are only effective on RGB images, 
our frameworks do not suffer from this limitation and 
are effective for arbitrary data.

\item[$\bullet$]\vspace{-0.3em}
Experiments on MNIST, depth images and RGB images, which validate our theoretical
finding and demonstrate the performance of our frameworks. 
We show that the proposed frameworks are not only theoretically optimal 
but also can practically achieve state-of-the-art perceptual quality. 
\end{itemize}\vspace{-0.8em}


It is worth noting that an optimal training framework for perfect perceptual decoding 
has been recently proposed in \cite{2021YanZy}, 
which only uses a single 
conditioned adversarial loss to train a perceptual decoder. Though theoretically 
optimal and has shown superiority on MNIST, it struggles to yield satisfactory 
performance on data with more complex distribution, e.g., RGB images. This is due to 
the fact that in practice adversarial loss only is insufficient for stable training 
of a decoder to decode RGB images from compressed representation. We augment the framework of 
\cite{2021YanZy} by additionally conditioning on an MMSE decoder. We prove that this 
augmentation does not compromise the optimality. Hence our frameworks are still optimal 
for training perfect perceptual decoder, and, more importantly, can achieve satisfactory 
performance on RGB images.


After completing this paper and when about to submit it for publication,
we became aware of \cite{2021AtheoryofDP}, who prove that,
for MSE distortion and Wasserstein-2 perception index,
optimal estimators on the D-P curve can be interpolated
from the estimators at the two extremes of the D-P curve, 
e.g., an MMSE estimator and a perfect perception estimator. 
This result is similar to our result in property 
\textit{ii)} of Theorem 1. However, our result is fundamentally 
different from that in \cite{2021AtheoryofDP} as follows: 
\textit{1)} Compared with the restoration problem considered in \cite{2021AtheoryofDP},
the lossy compression problem in this work requires 
jointly optimizing an encoder and a decoder, hence is more involved.
\textit{2)} Our analysis (proof of Theorem 1 \textit{ii)}) is 
fundamentally different from \cite{2021AtheoryofDP}
not only in that we have to jointly consider an encoder 
and a decoder in the D-P tradeoff formulation,
but also our analysis follows a completely different line, e.g., 
by analyzing the optimal solution of an unconstrained 
form of the D-P formulation (see Appendix A for details).
\textit{3)} Besides, we propose an augmented optimal formulation for perfect perceptual decoding learning and further propose two optimal training frameworks for D-P controllable lossy compression.

\section{Related Works}

\subsection{Perceptual Lossy Compression}

In learning based image compression, a common way to improve perceptual quality is to
incorporate an adversarial loss and/or a perceptual loss. Adversarial loss is typically 
implemented using generative adversarial networks (GAN) \cite{2014GAN}, 
whilst perceptual loss computes a distance on deep features 
\cite{2016Perceptual, 2015VeryDeep, 2016ImageStyle, 2017Photographic, 2016Generating}, 
e.g., MSE on the middle layers of the VGG net \cite{2015VeryDeep}.
Due to its high effectiveness in minimizing the divergence between distributions, 
adversarial learning has shown remarkable effectiveness in achieving high perceptual quality 
\cite{2017RealTime, 2019Extreme, 2017PhotoRealistic, 2019ESRGAN, 2020AGANbased, 2020HighFidelity}. 
Using a combination of distortion loss and adversarial loss,
the D-P tradeoff can be controlled by the balance parameter between these
two losses. However, simply combining distortion and adversarial losses is not optimal for
perceptual decoding and D-P tradeoff. Besides, it needs to train different models for 
different D-P tradoff. \cite{2021YanZy} proposes an optimal training framework which uses 
an MMSE encoder and train a perfect perceptual decoder using adversarial training.
Though can theoretically achieve the optimal condition of perfect perceptual decoding and has shown
superiority on MINIST data, only using adversarial loss cannot obtain satisfactory 
performance on data with more complex distribution, e.g., RGB images.




\subsection{Distortion-Perception Tradeoff}

While it is both theoretically and empirically demonstrated that 
distortion and perceptual quality are at odds with each other \cite{2018PDtradeoff, 2019Rethinking},
it is unclear how to optimally and flexibly achieve arbitrary D-P tradeoff.
The work \cite{2021YanZy} proves that the lossy compression problem with or
without perception constraint can share a same encoder, but it is limited 
to perfect perception constraint. Traditional methods using a combination 
of distortion loss, adversarial loss and deep features based loss can control
the D-P tradeoff by adjusting the balance parameters between the losses.
But for different perceptual quality (i.e. different D-P tradeoff), different 
encoder-decoder pairs need to be trained.
\cite{2021UniversalDP} further shows that the D-P tradeoff can be achieved 
by fixing an MMSE encoder and varying the decoder to (approximately) achieve 
any point along the D-P tradeoff. Even though it only requires to train a single encoder, 
different decoders need to be trained for different D-P tradeoff. 
\cite{2020Fidelity} considers the tradeoff between distortion and fidelity 
by interpolating the parameters or the outputs of two decoders. However, the method
is heuristic and not optimal.

\section{Theory for Optimally Controllable D-P Tradeoff}

\subsection{Perceptual Lossy Compression}

In lossy compression, the goal is to represent data with less bits and reconstruct data with low distortion \cite{2016Variable, 2017Soft2hard, 2017FullResolution, 2018Learning, 2018Conditional, 2018JointAutoregressive, 2021GuoZyStH} or high perceptual quality \cite{2017RealTime, 2019Extreme, 2017PhotoRealistic, 2019ESRGAN, 2020AGANbased, 2020Fidelity, 2020HighFidelity}. Recent studies have revealed that distortion and perceptual quality are at odds with each other. More specifically, high perceptual quality can only be achieved with some necessary increase of the lowest achievable distortion. Hence, a tradeoff between distortion and perception has to be considered \cite{2018PDtradeoff}. For the lossy compression problem, this has been recently taken into consideration by extending the classic rate-distortion tradeoff theory to the rate-distortion-perception three-way tradeoff \cite{2019Rethinking, 2018IntroducingPD2RD, 2018RDPtradeoff}.

The perceptual quality of a decoded sample refers to the degree to which it looks like a natural sample from human’s perception (subjective judgments), regardless its similarity to the reference source sample. A natural way to quantify perceptual quality in practice is to conduct real-versus-fake questionnaire studies \cite{2018Unreasonable, 2016Colorful, 2016ImprovedTech}. Conforming with this common practice, it can be mathematically defined in terms of the deviation between the statistics of decoded outputs and natural samples. For instance, denote the source and decoded output by $X$ and $\hat{X}$, respectively, the perceptual quality of $\hat{X}$ can be conveniently defined as \cite{2018PDtradeoff}
\begin{align}
d(p_X, p_{\hat{X}}),
\end{align}
where $d(\cdot, \cdot)$ is a deviation measure between two distributions, such as the KL divergence or Wasserstein distance. Such a definition correlates closely with human subjective score and accords well with the principles of no-reference image quality measures \cite{2013Making, 2005Reduced}.

With these understanding, a natural way to achieve perceptual decoding is to promote the decoded outputs to have a distribution as close as possible to that of natural samples. In practice, this can be implemented by incorporating an adversarial loss into training to use a distortion-plus-adversarial loss (DAL) as
\begin{align}
L= \lambda  L_{adv}+L_{dis},\label{preLfunc}
\end{align}
where $L_{dis}$ is a distortion loss such as MSE, $\ell_1$ norm, or distance between deep features, which computes the distortion between the reconstruction and its paired reference source. $L_{adv}$ is an adversarial loss responsible for minimizing the deviation from the distribution of natural samples to promote high perceptual quality. $ \lambda>0 $ is a balance parameter. Loss \eqref{preLfunc} and its many variants are widely used for achieving high perceptual quality and have shown remarkable effectiveness in fulfilling that goal \cite{2017RealTime, 2019Extreme, 2017PhotoRealistic, 2019ESRGAN, 2020AGANbased, 2020Fidelity, 2020HighFidelity}.

Though have shown good effectiveness, such methods have limitations as explained as follows. To achieve perfect perceptual quality, $d(p_X,p_{\hat{X}})\le 0$ (i.e. $p_X=p_{\hat{X}}$) is desired. Formulation \eqref{preLfunc} relaxes the constraint $d(p_X,p_{\hat{X}})\le 0$ by incorporating it with a distortion loss to get an unconstrained form. This relaxation eases the implementation, e.g., \eqref{preLfunc} can be implemented by a GAN and MSE loss in practice, but it cannot achieve perfect (or near perfect) perceptual quality with $d(p_X,p_{\hat{X}})\le 0$. Even though a very large value of $ \lambda $ would promote $d(p_X,p_{\hat{X}})$ to get close to 0, in this case the distortion cannot be well optimized and would result in undesirable excessive increase in distortion \cite{2021YanZy}. In Section 4, we propose two theoretically optimal training frameworks that do not suffer from such limitations.

With \eqref{preLfunc}, the distortion and perceptual quality of a compression system (i.e., D-P tradeoff) can be controlled by adjusting the value of $ \lambda $. Intuitively, to attain different D-P tradeoff, different models have to be trained for different values of $ \lambda $. Next, we show that only two specifically constructed decoders are enough for optimally achieving arbitrary D-P tradeoff.

\subsection{Theoretical Results for Optimally Controllable D-P Tradeoff}

When using the DAL framework \eqref{preLfunc} to achieve different D-P tradeoff, not only different models have to be trained but also it is not theoretically optimal. Here we present a theoretical result for optimal D-P tradeoff with only two decoders.

For a given bit-rate $R$, let $X$ be the source and $(E,G)$ be an encoder-decoder pair. When distortion is measured by MSE and perceptual quality is measured by squared Wasserstein-2 distance, the D-P tradeoff can be expressed as
\begin{align}
\begin{split}
D(P):=&\mathop {\min }\limits_{E\in \Omega,G} \ \mathbb{E} {\left\| {X - G(E(X))} \right\|^2} \\
&{\rm{s.t.}}~~~\ W^2_2({p_X},{p_{G(E(X))}}) \le P,\label{OriginProblem}
\end{split}
\end{align}\\
where $W^2_2(\cdot,\cdot)$ is squared Wasserstein-2 distance, and $\Omega$ is the set of encoders, of which the average bit-rate is $R$. 

When $P\!=\!+\infty $, \eqref{OriginProblem} degenerates to the classic MMSE formulation without considering perceptual quality, as the constraint is invalid in this case. Accordingly, $D(\infty)$ is the lowest achievable MSE. When $P\!=\!0$, the system is enforced to yield perfect perceptual quality, with $D(0)$ being the lowest achievable MSE under perfect perception constraint. 
$D(P)$ is nonincreasing and convex \cite{2018PDtradeoff}, which satisfies $D(0)\!=\!2D(+\infty)$ \cite{2021YanZy}. 

We now state the result that a linear interpolation between the outputs of 
two optimal decoders under \eqref{OriginProblem} with $P=+\infty$ and $P=0$, 
respectively, can achieve any points of the D-P tradeoff bound $D(P)$.

\hangafter=4
\setlength{\hangindent}{1em}
\textbf{Theorem 1.} 
\textit{Let $(E_d,G_d)$ be an optimal encoder-decoder pair to \eqref{OriginProblem} when $P\!=\!+\infty$,
and $G_p$ be an optimal decoder to \eqref{OriginProblem} for a fixed encoder $E_d$ and $P\!=\!0$. Denote $Z_d:=E_d(X)$ and $P_d:=W^{2}_{2}(p_X,p_{G_d(Z_d)})$. Then, these hold:\\
i) $E_d$ is an optimal encoder to \eqref{OriginProblem} for any $P \ge 0$.\\
ii) Let $\alpha=\min(\sqrt{P/P_d},1)\in[0,1]$, define
\[G^*_\alpha(Z_d) := \alpha G_d(Z_d)+(1-\alpha)G_p(Z_d),\]
then $(E_d, G^*_\alpha)$ is an optimal encoder-decoder pair to \eqref{OriginProblem}.}

When $P \ge P_d$, it is obvious that $\alpha=1$ and $(E_d,G^*_\alpha)=(E_d,G_d)$
is optimal to \eqref{OriginProblem} as it reaches the
lowest achievable MSE distortion $D(+\infty)=P_d=W^{2}_{2}(p_X,p_{G_d(Z_d)})$.
Meanwhile, from the definition that
$G_p$ is an optimal decoder to \eqref{OriginProblem}
for a fixed MMSE encoder $E_d$ when $P=0$,
and by the results in \cite{2021YanZy},
we have $\alpha=0$ and that $(E_d,G^*_\alpha)=(E_d,G_p)$ is optimal to \eqref{OriginProblem} when $P=0$.
Thus, to prove Theorem 1,
we only need to consider the case of $0\!<\!P\!<\!P_d$.
Consider an unconstrained formulation
corresponding to \eqref{OriginProblem} as
\begin{align}
\begin{split}
\mathop {\min }\limits_{E\in \Omega,G} \ &\alpha \mathbb{E} {\left\| X \!-\! G(E(X)) \right\|^2} + (1\!-\!\alpha) W^{2}_{2}({p_X},{p_{G(E(X))}}), \label{RelaxedW2Problem}
\end{split}
\end{align}
where $0<\alpha<1$.
To prove Theorem 1, we first prove that $(E_d, G^*_\alpha)$ is an optimal
encoder-decoder pair to \eqref{RelaxedW2Problem},
and then prove that, for any $0<P<P_d$,
$(E_d, G^*_\alpha)$ with $\alpha=\min(\sqrt{P/P_d},1)$ is also optimal to \eqref{OriginProblem}.
The details of proof are given in Appendix A.

\textbf{Remark 1.} 
\textit{Theorem 1 indicates that, for any $P\geq0$, 
an optimal decoder $G^*_\alpha$ to \eqref{OriginProblem} can be easily obtained 
by interpolating between an MMSE decoder $G_d$ and a perfect perceptual decoder $G_p$. 
Hence, it is unnecessary to train different decoders for different D-P tradeoff.
A simple linear interpolation between the outputs of the two decoders $G_d$ and 
$G_p$ is enough to reach any points of the D-P bound $D(P)$.
Furthermore, for an optimal encoder-decoder pair $(E_d,G^*_\alpha)$, 
the decoding distortion is
\begin{align}
\begin{split}
&\mathbb{E} {\|{X - (\alpha X_d + (1-\alpha) X_p)} \|^2}\\
\mathop {\rm{ = }}\limits^{(a)}&\mathbb{E} {\|{X_d - X} \|^2}+\mathbb{E} {\left\|{X_d - (\alpha X_d + (1-\alpha) X_p)} \right\|^2}\\
\mathop {\rm{ = }}\limits^{(b)}&[1+(1-\alpha)^2]\mathbb{E} {\left\|{X_d - X} \right\|^2}.\label{distortionD}
\end{split}
\end{align}
where $X_d=G_d(E_d(X))$ and $X_p=G_p(E_d(X))$. 
The proof of step (a) is given in Appendix A, 
and (b) is due to 
$\mathbb{E} {\|{X_d - X_p} \|^2}=\mathbb{E} {\|{X_d - X} \|^2}$ \cite{2021YanZy}. 
By varying the value of the interpolation factor $\alpha\in[0,1]$, 
we can control the tradeoff between the distortion and perceptual quality of decoding. 
When $\alpha=0$, the decoding distortion is exactly twice of 
that when $\alpha=1$, which is consistent with the result in \cite{2021YanZy}.}

\textbf{Remark 2.} 
\textit{Theorem 1 also implies that, an MMSE encoder is universal for
perceptual lossy compression in that it is optimal 
under any perception constraint, e.g., $\forall P\geq0$ in \eqref{OriginProblem}.
\cite{2021YanZy} has proved that when distortion is measured by MSE, 
an MMSE encoder is also optimal under perfect perception constraint 
(i.e. $P=0$ in \eqref{OriginProblem}). 
Our result expands this property to any $P\ge 0$, 
when perceptual quality is measured by squared Wasserstein-2 distance.}

\textbf{Remark 3.} 
\textit{From Theorem 1, we can further draw some properties on the D-P tradeoff function $D(P)$.
Specifically, when $P=+\infty$, we have
\begin{align}
\begin{split}
P_d=\mathop {\min }\limits_{p_{X,Z}} \mathbb{E} {\left\|{G_d(Z) - X} \right\|^2}=D_d,
\end{split}
\end{align}
where $D_d$ is the lowest achievable MSE distortion,
i.e., the distortion of an optimal MMSE encoder-decoder pair.
From \eqref{distortionD} and \eqref{squaredW2} in Appendix A, the relationship between distortion $D$ and perception $P$ is
\begin{align}
\begin{split}
&D=D_d+(1-\alpha)^2D_d,\\
&P=\alpha^2P_d=\alpha^2D_d. \label{DandP}
\end{split}
\end{align}
The first and second derivatives of $P$ with respect to $D$ are given by
\begin{align}
\begin{split}
\frac{dP}{dD}=&\frac{\alpha}{\alpha-1},\\
\frac{d^2P}{dD^2}=&\frac{1}{2(1-\alpha)^3D_d}. \label{2ndDerivative}
\end{split}
\end{align}
Then, it is easy to see that $\frac{dP}{dD}<0$ and $\frac{d^2P}{dD^2}>0$, $\forall\alpha\in(0,1)$. This is consistent with \cite{2019Rethinking} that the D-P tradeoff is nonincreasing and convex.}

It is worth noting from Theorem 1 that, to achieve interpolation based optimal D-P tradeoff, the MMSE decoder and the perfect perception decoder should be paired with a common MMSE encoder. Next we consider the construction of such decoders.

\section{Optimal Training Frameworks for Perfect Perceptual Decoding}

To realize the interpolation based D-P tradeoff as stated in Theorem 1,
an MMSE decoder $G_d$ and a perfect perceptual decoder $G_p$ (both paired with a common MMSE encoder $E_d$) are needed.
The encoder-decoder pair $(E_d,G_d)$ can be straightforwardly obtained via MMSE training.
Meanwhile, the decoder $G_p$ can be trained by the framework in \cite{2021YanZy},
which is theoretically optimal for perfect perceptual decoding. 
It is proved that the mapping of an optimal encoder-decoder pair with perfect perceptual quality is symmetric. 
The optimal condition is that the source $X$ and decoded output $\hat{X}$ have a same  distribution conditioned on the compressed representation $Z$ as
\begin{align}
{p_{\hat{X}|Z}} = {p_{X|Z}}.\label{theorem1}
\end{align}

This can be fulfilled by a two-stage training procedure \cite{2021YanZy}:
firstly train an encoder-decoder pair $(E_d,G_d)$ via minimizing MSE as
\begin{align}
\mathop {\min }\limits_{E\in \Omega,G} \mathbb{E} \left\|X-G(E(X))\right\|^2, \label{ConditionAdvTraining1}
\end{align}
then fix the encoder $E_d$ and train a perceptual decoder $G_p$ by 
\begin{align}
\mathop {\min }\limits_{{p_{\hat X,Z_d}}} W_1(p_{\hat X,Z_d},p_{X,Z_d}),\label{Stage2ofYan}
\end{align}
where $Z_d:=E_d(X)$, $W_1$ is the Wasserstein-1 distance, which is implemented by WGAN with
a conditional discriminator \cite{2014ConGAN} to discriminate between $(\hat{X},Z_d)$ and $(X,Z_d)$. 

Since distortion loss is not used in training $G_p$, 
the practical performance depends heavily on adversarial training. 
Although this framework has shown superiority over the DAL framework on MNIST, 
intensive experiments show that adversarial loss only (e.g. \eqref{Stage2ofYan}) 
cannot yield satisfactory performance on RGB images.



\subsection{An Augmented Optimal Formulation for Perfect Perceptual Decoding Learning}

To address the limitation of the above framework,
here we augment it to achieve satisfactory practical performance on RGB images,
but without compromising its optimality.
We propose an augmented formulation of \eqref{Stage2ofYan} as
\begin{align}
\mathop {\min }\limits_{{p_{\hat X,X_d}}} W_1(p_{\hat X,X_d},p_{X,X_d})+ \lambda \mathbb{E}\|\hat{X}-X_d \|,\label{normOFun}
\end{align}
where $X_d:=G_d(E_d(X))$ with $(E_d,G_d)$ being a given MMSE encoder-decoder pair,
$\lambda$ is a positive parameter associated with the augmentation term.
Intuitively, formulation \eqref{normOFun} incorporates an additional supervision provided by the $\ell_2$ distance from the MMSE decoding $X_d$, which is expected to assist the adversarial training.

Next, we prove that this augmentation does not change 
the optimal solution of \eqref{Stage2ofYan},
i.e., it is still optimal for perfect perceptual decoding learning. 
The proof is given in Appendix B.


\textbf{Theorem 2.} 
\textit{Let $(E_d,G_d)$ be an MMSE encoder-decoder pair, and $W_1(\cdot,\cdot)$ be the Wasserstein-1 distance. Denote $Z_d=E_d(X)$ and $X_d=G_d(Z_d)$, then these hold:
\begin{itemize}	
\item[i)]\vspace{-1em}
When $0 \le \lambda<1$, the optimal solution of \eqref{normOFun} satisfies 
$p_{\hat X,X_d}=p_{X,X_d}$, or equivalently $p_{\hat X,Z_d}=p_{X,Z_d}$.
\item[ii)]\vspace{-0.4em}
When $\lambda>1$, the optimal solution of \eqref{normOFun} satisfies $\hat{X}=X_d$.
\end{itemize}}

\textbf{Remark 4.} 
\textit{Theorem 2 implies that, with any $\lambda\in[0,1)$,  formulation \eqref{normOFun} can achieve the optimal condition $p_{\hat X,Z_d}=p_{X,Z_d}$ (equivalently $p_{\hat X,X_d}=p_{X,X_d}$) for perfect perception decoding. Hence, it is still optimal for perfect perception decoding.
In practice, the augmentation term is very helpful to enhance the training of the perceptual decoder, as will be shown in experiments. This benefit is mainly thanks to the fact that the augmentation term, which provides strong supervision by the $\ell_2$ distance from the MMSE decoding $X_d$, can effectively stabilize the training procedure.
}

\textbf{Remark 5.} 
\textit{Theorem 2 provides a solid theoretical foundation 
for the augmented formulation \eqref{normOFun} that 
its solution satisfies the optimal condition for any $\lambda\in[0,1)$. 
However, to yield satisfactory performance in practical applications, 
the value of $\lambda$ still needs to be tuned. 
This is due to the facts that: 
1) in practical implementation the Wasserstein-1 distance can only be approximated,
e.g., by WGAN which involves 1-Lipschitz approximation \cite{2017Wasserstein};
2) the capacity of a decoding model (e.g., a deep network) in practice is not infinite.
Thus, different values of $\lambda\in[0,1)$ would yield different practical performance, 
as illustrated in Table 2 in Appendix D. 
Another empirical illustration of Theorem 2 on MNIST is given in Figure 2.
}

\subsection{Proposed Optimal Training Frameworks}

\begin{figure*}[!t]
	\vskip 0.2in
	\begin{center}
		\subfigure[Framework A]{
			\begin{minipage}[t]{0.9\columnwidth}
				\label{2decoders}
				\centering
				\includegraphics[width=\columnwidth]{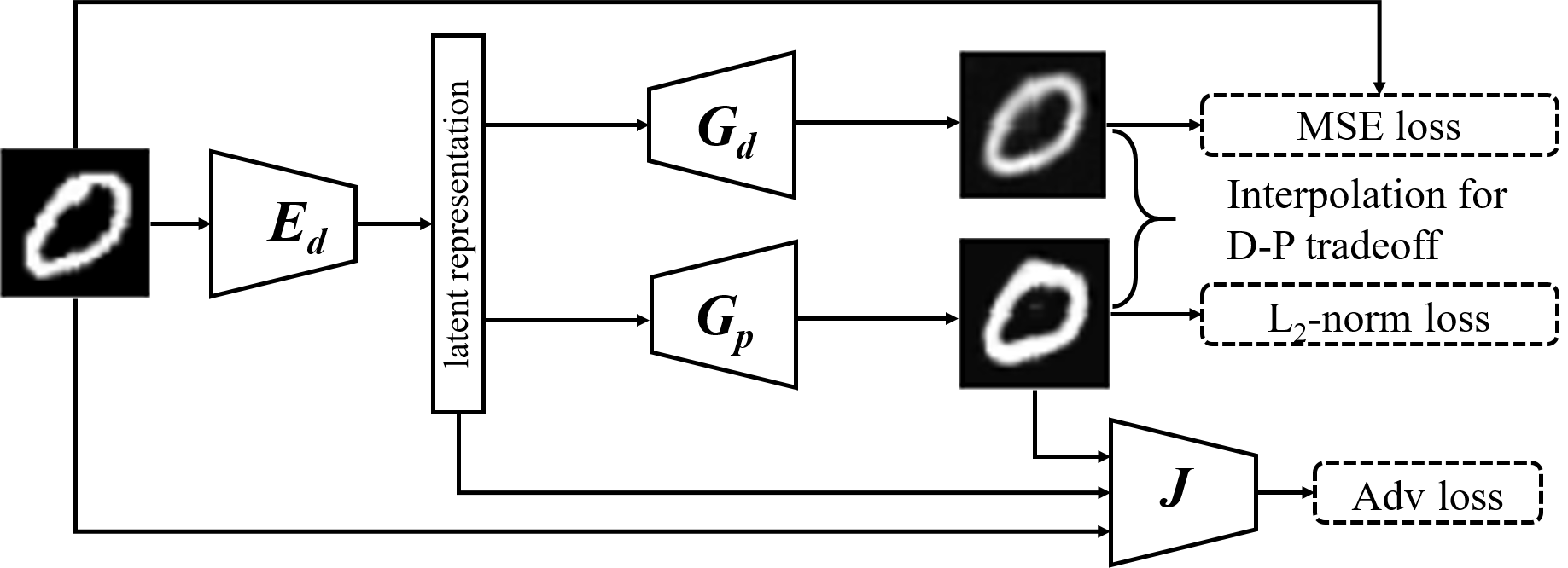}
			\end{minipage}%
		}%
		\hspace{0.1\columnwidth}
		\subfigure[Framework B]{
			\begin{minipage}[t]{0.9\columnwidth}
				\label{postprocess}
				\centering
				\includegraphics[width=\columnwidth]{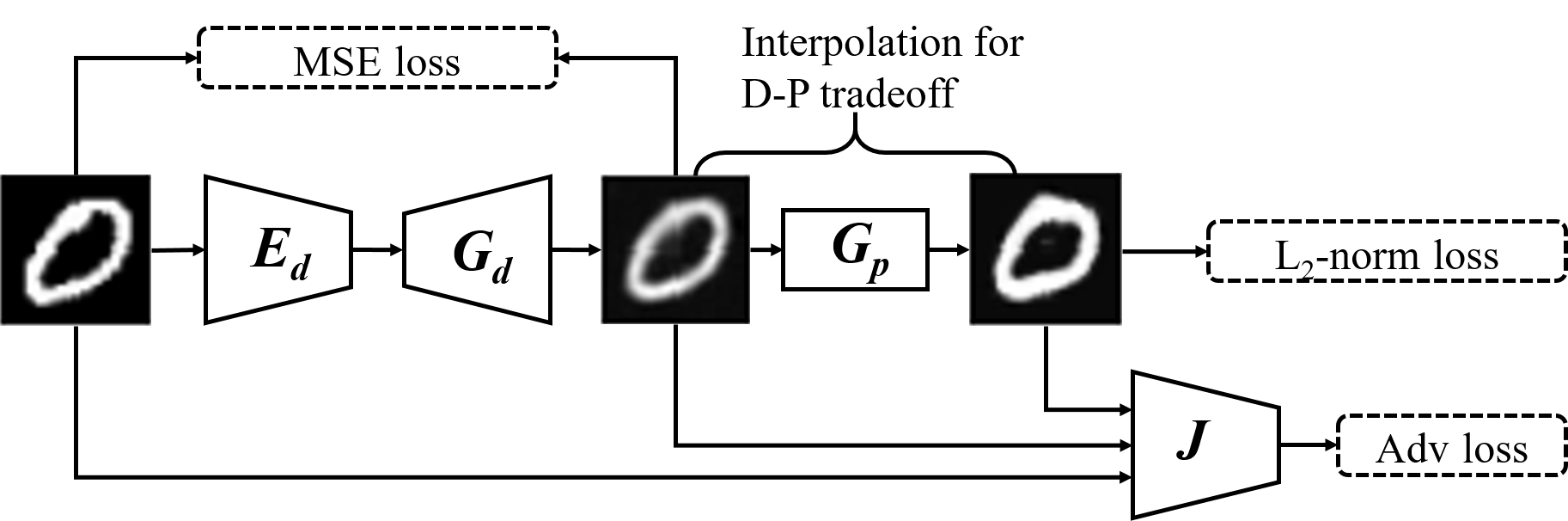}
			\end{minipage}%
		}
		
	\caption{Proposed optimal training frameworks for perfect perceptual decoding which applies to interpolation based D-P tradeoff.}
	\end{center}
	\vskip -0.2in
\end{figure*}

Based on the above theoretical results,
we propose two frameworks for perfect perceptual decoding training,
which have different characteristics that may be discriminatively preferred in different applications.

The first is illustrated in Figure~\ref{2decoders},
which consists an MMSE encoder-decoder pair $(E_d,G_d)$ and
a perfect perceptual decoder $G_p$ paired with $E_d$.
The two decoders are separately trained by two steps:
firstly train $(E_d,G_d)$ by the MSE loss,
then fix the encoder to $E_d$ and train $G_p$ via \eqref{normOFun}.
In the second step, \eqref{normOFun} can be implemented by WGAN-gp \cite{2017Wasserstein},
with the $\ell_2$ loss being incorporated into the adversarial loss of WGAN-gp
to train $G_p$ and a discriminator $J$ alternatively.
From Theorem 2, when using $\lambda<1$,
the optimal condition for perfect perceptual decoding would be reached
when $(G_p,J)$ are trained to be optimal.

The second is illustrated in Figure~\ref{postprocess},
which has a combination structure consists of an MMSE
encoder-decoder pair $(E_d,G_d)$ followed by a
post-processing perceptual decoder $G_p$.
It also involves two steps that firstly train an MMSE encoder-decoder pair $(E_d,G_d)$
and then train $G_p$ for post perceptual mapping from the output of $G_d$.
That is $G_p$ is also trained via \eqref{normOFun} but with $\hat{X}=G_p(G_d(E_d(X)))$.

In theory, both frameworks can optimally achieve
arbitrary points of the D-P bound \eqref{OriginProblem}
by simply interpolating between $G_d$ and $G_p$
according to Theorem 1.
While Framework A is straightforward and has a more compact structure,
Framework B is more flexible that applies to any existing lossy 
compression system that optimized only in terms of decoding distortion.
As will be shown in experiments (Appendix C),
the MMSE encoder-decoder pair $(E_d,G_d)$ in Framework B
can be replaced by an compression system only optimizing 
decoding distortion, e.g., BPG.

Note that in \eqref{OriginProblem} and Theorem 1,
the perceptual quality is measured by squared 
Wasserstein-2 distance, while here we use Wasserstein-1 
distance for perfect perceptual decoding training (see \eqref{normOFun}).
This does not affect the optimality of the proposed frameworks
when used for interpolation based D-P tradeoff,
since Theorem 1 only requires $G_p$ to satisfy the optimal condition
$P_{\hat{X},Z_d}\!=\!P_{{X},Z_d}$ (equivalently $P_{\hat{X},X_d}\!=\!P_{{X},X_d}$),
regardless it is achieved by Wasserstein-1 distance, Wasserstein-2 distance, or any other divergence.



\section{Experiments}

We provide experiment results on MNIST, depth and RGB images. 
First, we verify Theorem 2 on MNIST \cite{1998Gradient} by examining
the effect of $\lambda$ in \eqref{normOFun}.
Then, we evaluate the proposed frameworks on depth and RGB images.
For both Framework A and B, $E_d$, $G_d$ are set the same as HiFiC \cite{2020HighFidelity}. For framework A, the generator $G_p$ uses a convolutional layer 
to convert the encoded representation into a matrix with the same shape 
as the input images. The rest structure is set the same as the 
post-processing decoder in Framework B.
For Framework B, the generator (post-processing decoder) has an 
U-net architecture, which consists of two down-sampling layers and two up-sampling 
layers with the use of residual channel attention block (RCAB) \cite{2018RCAB, 2022OT4Denoise}.
Skip connection is used to learn the residual.
The discriminator has one more down-sampling layer than that in \cite{2017PhotoRealistic} 
because the patch size is $128\times 128$.




\subsection{Results on MNIST}

\begin{figure}[!t]
	\vskip 0.0in
	\begin{center}
		\centerline{\includegraphics[width=0.9\columnwidth]{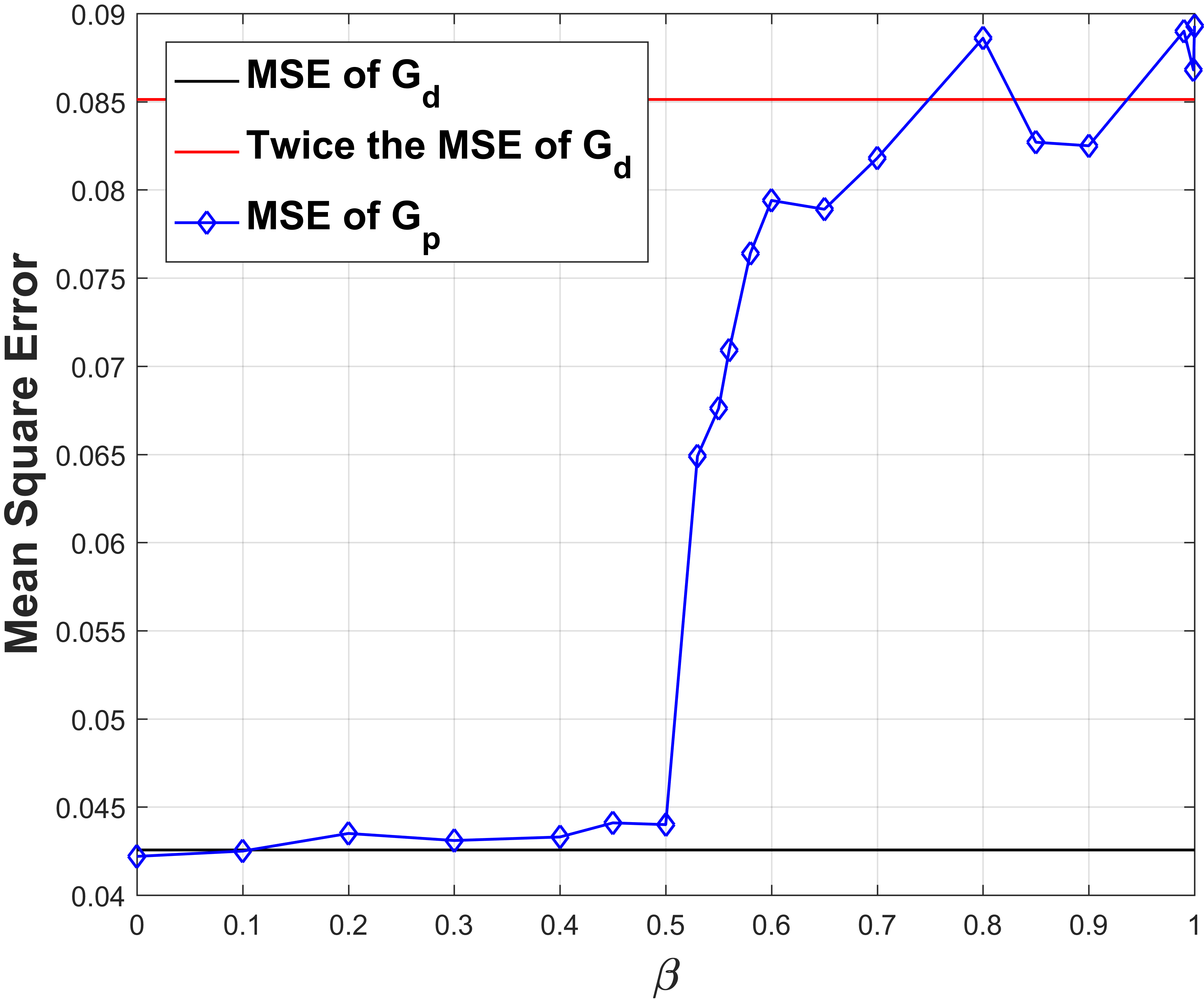}}
		\vskip -0.1in
		\caption{Empirical decoding distortion of Framework A versus $\beta$ ($\lambda$ in \eqref{normOFun} is related to $\beta$ in \eqref{mnistloss} as $\lambda=\frac{1-\beta}{\beta}$).}
		\label{distortion-alpha}
	\end{center}
	\vskip -0.4in
\end{figure}

\begin{figure*}[!t]
	\vskip 0.2in
	\begin{center}
		\subfigure[Results of the traditional DAL framework \eqref{preLfunc} with different values of $\lambda$.]
		{
			\begin{minipage}[t]{1.0\columnwidth}
				\centering
				\includegraphics[width=\columnwidth]{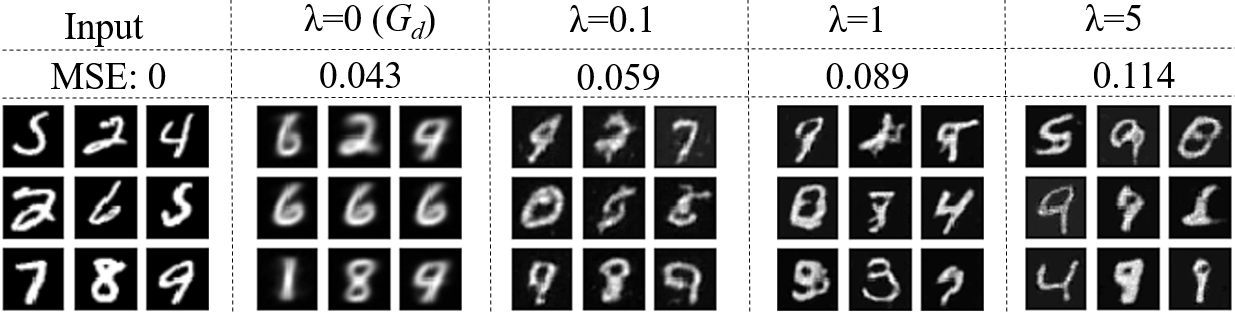}
			\end{minipage}%
		}\hspace{2mm}%
		\subfigure[Results interpolated between $G_d$ and $G_p$, i.e., $X_\alpha$ in \eqref{dptradeoff}. $G_p$ is trained by the proposed Framework A.]
		{
			\begin{minipage}[t]{1.0\columnwidth}
				\centering
				\includegraphics[width=\columnwidth]{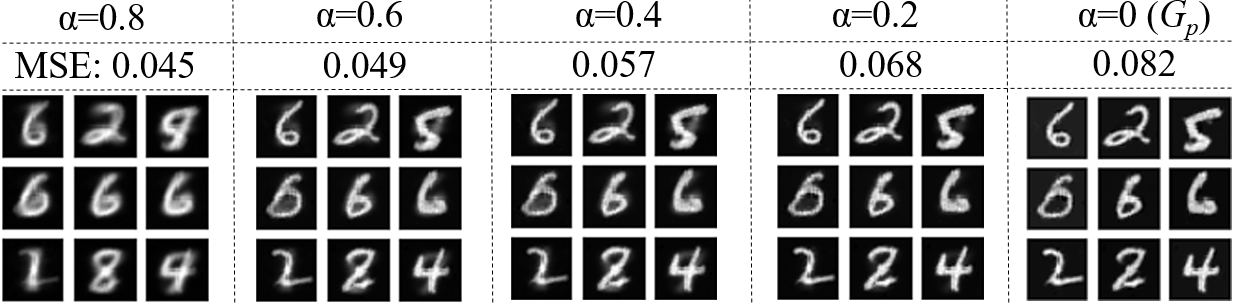}
			\end{minipage}%
		}%
		\caption{Typical sample comparison between the traditional DAL framework \eqref{preLfunc} and the proposed Framework A on MNIST.}
		\label{MNISTimage}
	\end{center}
	\vskip -0.2in
\end{figure*}

\begin{figure*}[!t]
	\vskip 0.2in
	\begin{center}
		\subfigure[Ground-truth (depth and its corresponding point cloud)]{
			\begin{minipage}[t]{0.515\columnwidth}
				\centering
				\includegraphics[width=\columnwidth]{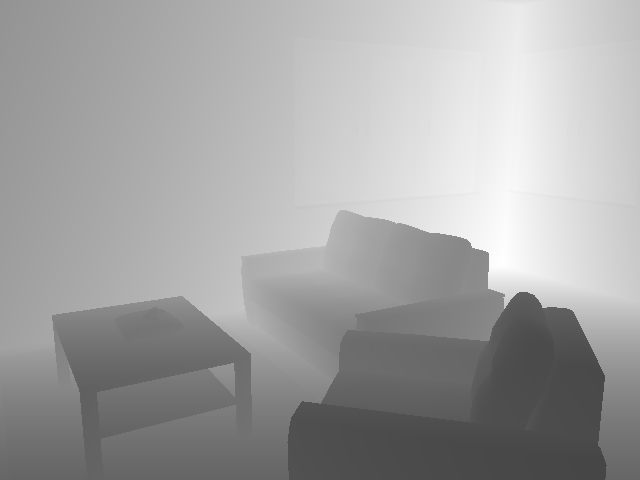}
			\end{minipage}~%
			\begin{minipage}[t]{0.4\columnwidth}
				\centering
				\includegraphics[width=\columnwidth]{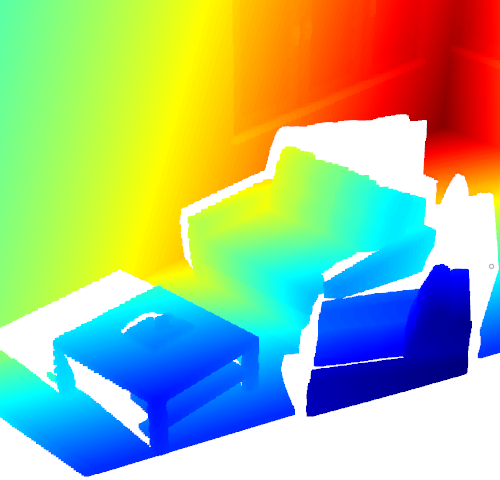}
			\end{minipage}%
		}~~~%
		\subfigure[$G_d$ (MMSE)]{
			\begin{minipage}[t]{0.4\columnwidth}
				\centering
				\includegraphics[width=\columnwidth]{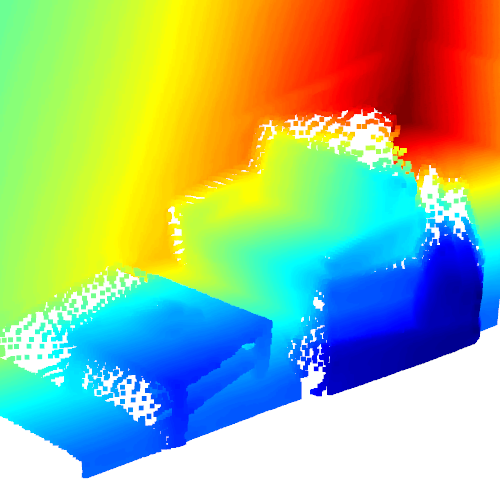}
			\end{minipage}%
		}~~~%
		\subfigure[$G_h$ (HiFiC)]{
			\begin{minipage}[t]{0.4\columnwidth}
				\centering
				\includegraphics[width=\columnwidth]{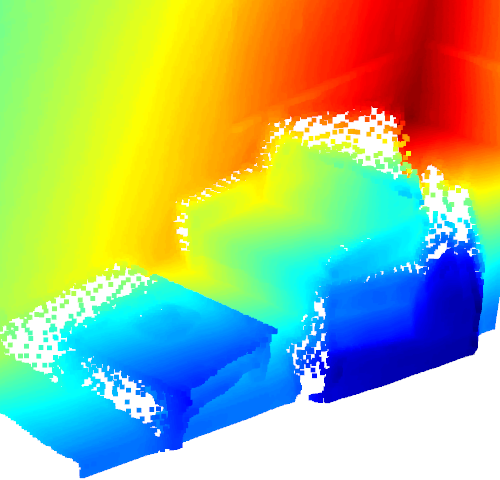}
			\end{minipage}%
		}%
		
		\subfigure[$G_p$ ($\alpha$=0)]{
			\begin{minipage}[t]{0.4\columnwidth}
				\centering
				\includegraphics[width=\columnwidth]{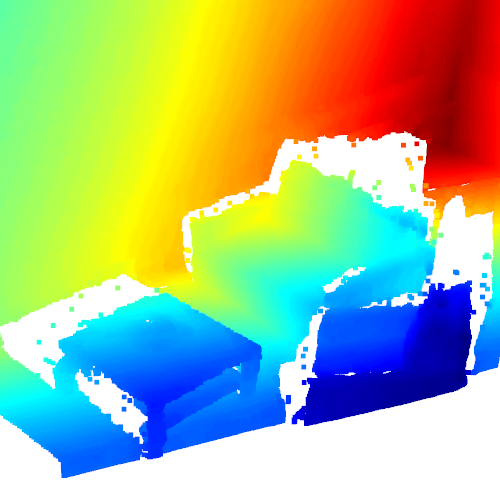}
			\end{minipage}%
			\label{suncg-d}
		}~~~%
		\subfigure[$\alpha$=0.25]{
			\begin{minipage}[t]{0.4\columnwidth}
				\centering
				\includegraphics[width=\columnwidth]{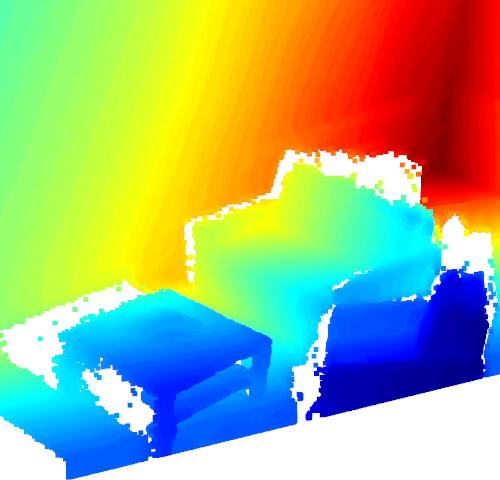}
			\end{minipage}%
		}~~~%
		\subfigure[$\alpha$=0.5]{
			\begin{minipage}[t]{0.4\columnwidth}
				\centering
				\includegraphics[width=\columnwidth]{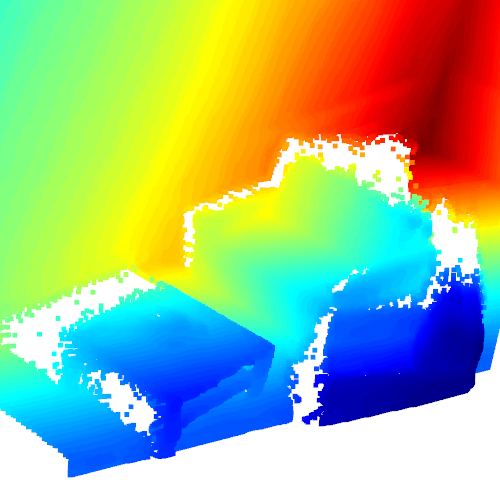}
			\end{minipage}%
		}~~~%
		\subfigure[$\alpha$=0.75]{
			\begin{minipage}[t]{0.4\columnwidth}
				\centering
				\includegraphics[width=\columnwidth]{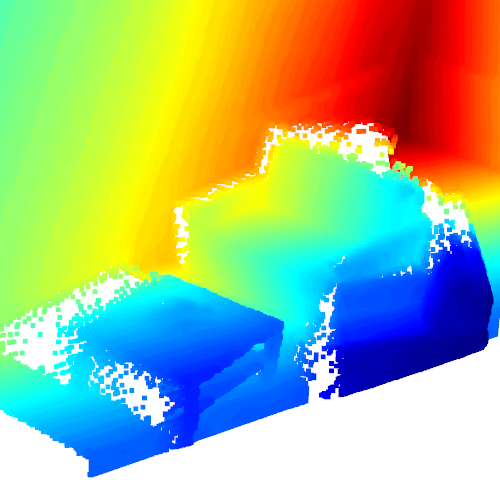}
			\end{minipage}%
		}%
\caption{Results on a sample from SUNCG dataset. (a) Ground-truth depth image and the corresponding point cloud. (b)-(d) Point cloud reconstructed from $G_d$, HiFiC and $G_p$. (e)-(g)  Interpolated results between $G_d$ and $G_p$. For visual clarity, point cloud results are shown.}
		\label{suncg}
	\end{center}
	\vskip -0.2in
\end{figure*}

We train 20 compression models by Framework A for a bit-rate of 4,
with the networks same as that in \cite{2021YanZy}. 
For a fixed encoder 
$E_d$ trained by MMSE, we train the decoder $G_p$ using WGAN-gp 
\cite{2017Wasserstein} by
\begin{align}
\begin{split}
&\mathop {\max }\limits_{\left\| J \right\|_L \le 1}  \mathbb{E}[J(G_p(E(X)),E(X))] - \mathbb{E}[J(X,E(X))]\\
&\mathop {\min }\limits_{{G_p}} (1-\beta) \mathbb{E} {\left\| {G_p(X) - X_d} \right\|}- \beta \mathbb{E}[J(G_p(X),E(X))], \label{mnistloss}
\end{split}
\end{align}
where $J$ is a discriminator lies within the space of 1-Lipschitz functions, 
$X_d$ is the output of an MMSE encoder-decoder pair, 
$\beta$ is a balance parameter related to $\lambda$ in 
\eqref{normOFun} as $\lambda=\frac{1-\beta}{\beta}$. 
Theorem 2 implies that the MSE of $G_p$ is the same as 
$G_d$ when $\beta < 0.5$, but would double when $\beta > 0.5$. 
Thus, in theory there would be a jump of the MSE
at $\beta = 0.5$ from $D(+\infty)$ (MSE of $G_d$) to $2D(+\infty)$ 
(double the MSE of $G_d$).

Figure~\ref{distortion-alpha} shows the empirical MSE of Framework A versus $\beta$.
The theoretical jumping property can be (approximately) observed.
As explained in Remark 5, the practical performance is related to the value of $\lambda$.
Figure~\ref{MNISTimage} shows the D-P tradeoff by interpolating the outputs of 
$G_d$ and $G_p$ as
\begin{align}
X_{\alpha}={\alpha}G_d(E_d(X))+(1-\alpha)G_p(E_d(X)).\label{dptradeoff}
\end{align}
As expected, as $\alpha$ varying from 0 to 1, 
the decoding output becomes more blurry but has lower distortion.
In Figure~\ref{MNISTimage}, the traditional DAL framework \eqref{preLfunc} 
(with different $\lambda$) is also compared. Clearly, for a similar 
MSE distortion result, the output of our framework is more clear.

\subsection{Results on Depth Images}

We further provide results on the SUNCG dataset \cite{2017SUNCG}. 
For depth image compression, we use the post-processing framework, 
i.e. Framework B as shown in Figure~\ref{postprocess}. 
We first pretrain an encoder-decoder pair $(E_d,G_d)$ 
for 100 epochs by MMSE with regime set as `low'. 
Then our generator $G_p$ is trained by \eqref{normOFun} with $\lambda=0.005$. 
For comparison, we train the HiFiC model for 100 epochs using 
the same hyper-parameters as in \cite{2020HighFidelity}, denoted by $G_h$. 
Figure~\ref{suncg} shows the decoding results on a depth image. 
For visual clarity, we show the point cloud results reconstructed from the depth results. 
For MMSE decoding, the edges are blurry, resulting in a large amount of 
noisy points around edges in the point cloud result. Besides, the output of $G_h$ contains many noisy points around edges, 
which is even not distinctly better than that of the MMSE decoding.
The perceptual quality of $G_p$ is 
much better, as shown in Figure~\ref{suncg-d}. 
This is thanks to that the generator can well learn the 
distribution of clean depth images. 
For the interpolated results by \eqref{dptradeoff}, 
as $\alpha$ increases, 
the quality of point cloud degrades with the edges becoming more blurry.



As the proposed Framework B applies to any existing compression system
that optimized only under certain distortion measure,
an illustration of applying it to post-process the BPG codec
is provided in Appendix C.

\subsection{Results on RGB Images}



\begin{table}[t]
\caption{Perception score comparison on the KODAK dataset in two bit-rate conditions.}
\label{sample-table}
\vskip 0.15in
\begin{center}
\begin{small}
\begin{sc}
\begin{tabular}{lcccr}
\toprule
 & \multicolumn{2}{c}{Low (0.13 bpp)} & \multicolumn{2}{c}{High (0.41 bpp)} \\
Methods & PSNR & PI & PSNR & PI \\
\midrule
GT    & $\infty$ & 2.23& $\infty$ & 2.23 \\
$G_d$ (MMSE) & 38.36& 4.54 & 40.22 & 3.11\\
$G_h$ (HiFiC)    & 37.94 & 2.10 & 39.98 & 2.24\\
\midrule
Framework A & & & &\\
$\alpha=0$ ($G_p$)    & 36.89 & 2.05 & 38.71 & 2.16\\
$\alpha=0.25$    & 37.45 & 2.18 & 39.28 & 2.15\\
$\alpha=0.50$    & 37.94 & 2.47 & 39.78 & 2.29\\
$\alpha=0.75$    & 38.27 & 3.32 & 40.13 & 2.73\\
\midrule
Framework B & & & &\\
$\alpha=0$ ($G_p$)    & 36.87 & 2.08 & 38.60 & 2.18\\
$\alpha=0.25$    & 37.42 & 2.20 & 39.18 & 2.18\\
$\alpha=0.50$    & 37.90 & 2.48 & 39.71 & 2.32\\
$\alpha=0.75$    & 38.24 & 3.33 & 40.09 & 2.73\\
\bottomrule
\label{perceptiontable}
\end{tabular}
\end{sc}
\end{small}
\end{center}
\vskip -0.1in
\end{table}

%
%

\begin{figure}[!t]
	\vskip 0.0in
	\begin{center}
		\centerline{\includegraphics[width=1.1\columnwidth]{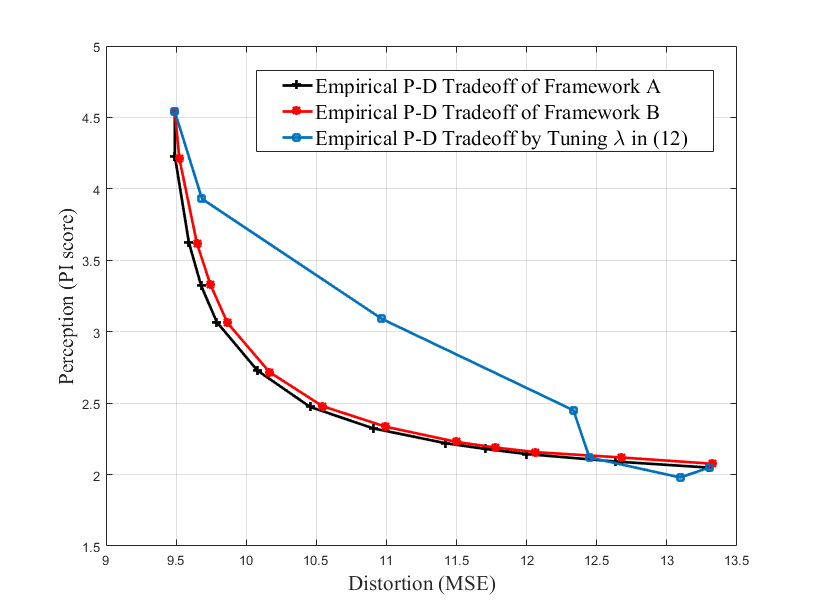}}
		\vskip -0.1in
		\caption{Empirical results on interpolation based perception-distortion tradeoff on the KODAK dataset.}
		\label{P-Dcurve}
	\end{center}
	\vskip -0.3in
\end{figure}

In the experiment on RGB images, the networks are set the same as the depth image experiment, except the input channel number is 3 and $\lambda=0.01$. 
We train $G_d$, $G_p$, $G_h$ on COCO2014 \cite{2014coco} in two different bit-rates, with regime set as `low' and `high', and 
interpolate the outputs of $G_d$ and $G_p$ by \eqref{dptradeoff} for D-P tradeoff. 
The perceptual quality is evaluated in terms of PI score \cite{2018pirm}, which is calculated by NIQE \cite{2013NIQE} and a learned network \cite{2017ChaoMa}. 
As shown in Table~\ref{perceptiontable}, the perception scores of $G_p$ outperforms 
HiFiC $G_h$ but with higher distortion. For each the proposed frameworks,
as $\alpha$ increases, the distortion decreases but the PI score deteriorates. 
However, the perception score of HiFiC is better than the interpolated results 
of both the new frameworks when $\alpha=0.5$, with similar distortion. 
This is because our theorem result is derived on squared Wasserstein-2 distance. 
We also plot the empirical perception-distortion curve 
in Figure~\ref{P-Dcurve}, including interpolated results by the two proposed frameworks, in comparison with the perception-distortion curve yielded by tuning $\lambda$ in \eqref{normOFun}. Since the samples in KODAK contains only 24 images, 
we use the PI score as perception measure rather than squared Wasserstein-2 distance. 
The perception-distortion curve of Framework A is similar to that of Framework B.

RGB samples for qualitative visual comparison are provided in Appendix D, see Figure~\ref{KODAK-1}-\ref{KODAK-3}.
Moreover, both of the proposed frameworks outperform the networks trained by directly altering $\lambda$. The experimental details and the effect of the parameter $\lambda$ on the practical performance of Framework A is evaluated in Appendix D, see Table \ref{lambdatable} and Figure~\ref{lambdasample} in Appendix D.


\section{Conclusion}

We presented a theoretical finding for the D-P tradeoff
problem in lossy compression that, two specifically constructed 
decoders are enough for arbitrary D-P tradeoff. 
We proved that arbitrary points of the D-P tradeoff bound can be 
achieved by simply interpolating between an MMSE decoder and a 
perfect perceptual decoder. Furthermore, we proposed two theoretically
optimal training frameworks for perfect perceptual decoding.
Experiments on MNIST, depth images and RGB images well verified our 
theoretical results and demonstrated the effectiveness of 
the proposed frameworks.

Taken together, this work lays a theoretical foundation for
not only simply achieving optimal D-P tradeoff in lossy compression
but constructing optimal training frameworks for perfect perceptual 
decoding that goes beyond existing heuristic frameworks.

\section*{Acknowledgement}

This work was supported in part by the National Natural Science Foundation of China (NSFC) under Grant 61871265 and the Science and Technology Innovation 2030-Major Project (Brain Science and Brain-Like Intelligence Technology) under Grant 2022ZD0208700.

\nocite{langley00}

\bibliography{reference_paper}
\bibliographystyle{icml2022}

\newpage
\appendix
\onecolumn
\section{Proof of Theorem 1.}

This section gives the proof of Theorem 1. When $P \ge P_d$, 
it is obvious that $\alpha=1$ and $(E_d,G^*_\alpha)=(E_d,G_d)$ 
is optimal to \eqref{OriginProblem} as it reaches the lowest 
achievable MSE distortion $D(+\infty)=P_d=W^{2}_{2}(p_X,p_{G_d(Z_d)})$. 
Meanwhile, since $G_p$ is an optimal decoder to \eqref{OriginProblem}
for a fixed MMSE encoder $E_d$ when $P=0$, by the results in \cite{2021YanZy}
we have $\alpha=0$ and that $(E_d,G^*_\alpha)=(E_d,G_p)$ is optimal 
to \eqref{OriginProblem} when $P=0$.
Thus, to prove Theorem 1, we only need to consider the case of $0<P<P_d$. 
Next, we first prove that $(E_d,G^*_\alpha)$ is an optimal encoder-decoder pair to \eqref{RelaxedW2Problem}, which is an unconstrained formulation of \eqref{OriginProblem}. 
Then, we prove that $(E_d,G^*_\alpha)$ is also an optimal encoder-decoder pair to \eqref{OriginProblem} by analyzing the relationship between the optimal solutions of \eqref{OriginProblem} and \eqref{RelaxedW2Problem}.


Denote $Z:=E(X)$, and consider an MMSE decoder $G_1(Z):=\mathbb{E}[X|Z]$, the first term of \eqref{RelaxedW2Problem} can be expressed as
\begin{align}
\begin{split}
\mathbb{E} {\left\| {X - G(Z)} \right\|^2}
=&\mathbb{E} {\left\| {X  - G_1(Z) + G_1(Z) - G(Z)} \right\|^2}\\
=&\mathbb{E} {\left\| {X - G_1(Z)} \right\|^2} + \mathbb{E} {\left\| {X_1 - G(Z)} \right\|^2}\\& + 2\mathbb{E} {\left \langle X-G_1(Z), X_1-G(Z)\right \rangle}\\
\mathop {\rm{ = }}\limits^{(a)}&\mathbb{E} {\left\| {X - G_1(Z)} \right\|^2} + \mathbb{E} {\left\| {G_1(Z) - G(Z)} \right\|^2}, \label{ExpandofDis}
\end{split}
\end{align}
where $<\cdot,\cdot>$ denotes the inner product, and $(a)$ is due to 
\begin{align}
\begin{split}
&\mathbb{E} {\left \langle X-G_1(Z), G_1(Z)-G(Z)\right \rangle}\\
=&\mathbb{E} \{\mathbb{E}[\left \langle X-G_1(Z),G_1(Z)-G(Z)\right \rangle|G_1(Z)]\}\\
=&\mathbb{E} \{ \mathbb{E}[(X-G_1(Z))|G_1(Z)]\cdot \mathbb{E}[G_1(Z)-G(Z)|G_1(Z)]\}\\
=&\mathbb{E}\{ 0\cdot \mathbb{E}[G_1(Z)-G(Z)|G_1(Z)]\}=0.
\end{split}
\end{align}

Now, we first find an optimal decoder to \eqref{RelaxedW2Problem} for a fixed encoder $E:X\to Z$, of which the joint distribution is denoted by $p_{X,Z}$. Denote $\hat{X}:=G(Z)$, the decoding mapping $G$ in \eqref{RelaxedW2Problem} can be expressed as $p_{Z, \hat{X}}$. To avoid symbol confusion, we consider an auxiliary variable $X'$ which has the same distribution as $X$, i.e. $p_{X'}=p_X$, hence $W^2_2(p_{X'},p_{\hat{X}})=W^2_2(p_{X},p_{\hat{X}})$. 
Then, from \eqref{ExpandofDis} and the fact that $\mathbb{E} {\left\| {X - G_1(Z)} \right\|^2}$ only depends on the encoder $E$, formulation \eqref{RelaxedW2Problem} with a fixed encoder $E$ can be written as
\begin{align}
\begin{split}
&\mathop {\min }\limits_{p_{Z, \hat{X}}} \ \alpha \mathbb{E} {\| {\hat{X} - G_1(Z)} \|^2} + (1-\alpha) W^{2}_{2}({p_{X'}},{p_{\hat{X}}})\\
=&\mathop {\min }\limits_{p_{Z,\hat{X}}} \alpha \mathbb{E} {\| {\hat{X}-G_1(Z)} \|^2} + \mathop {\min }\limits_{p_{X',\hat{X}}} \ (1-\alpha) \mathbb{E} {\| {X' - \hat{X}} \|^2} \\
=&\mathop {\min }\limits_{p_{X',Z,\hat{X}}} \alpha \mathbb{E} {\| {\hat{X}-G_1(Z)} \|^2} + (1-\alpha) \mathbb{E} {\| {X' - \hat{X}} \|^2} \\
=&\mathop {\min }\limits_{p_{X',Z}} \mathop {\min }\limits_{p_{\hat{X}|X,Z}} \ \alpha \mathbb{E} {\| {\hat{X}-G_1(Z)} \|^2} + (1-\alpha) \mathbb{E} {\| {X' - \hat{X}} \|^2},\label{ObjFofW2Problem}
\end{split}
\end{align}
where $X'$ is constrained by $P_{X'}=P_X$.
To find the optimal mapping $Z\to\hat{X}$ of \eqref{ObjFofW2Problem}, we first find the optimal solution of $\hat{X}$ when fixing $Z$ and $X'$, 
and then find the distribution of $X'$ conditioned on $Z$. 
Firstly, given $Z$ and $X'$, it is easy to see that 
any optimal $\hat{X}$ to \eqref{ObjFofW2Problem},
denoted by $\hat{X}^*$, satisfies
\begin{align}
\mathbb{E}[2\alpha(\hat{X}^*-G_1(Z)) + 2(1-\alpha)(\hat{X}^*-X')]=0. \label{DivofW2Problem}
\end{align}
Therefore, given $Z$ and $X'$, the optimal $p_{\hat{X}|X',Z}$ to \eqref{ObjFofW2Problem} satisfies 
\begin{equation}
    \hat{X}^*=\alpha G_1(Z) + (1-\alpha)X'. \label{OptimalXhat}
\end{equation}
Then, substituting $\hat{X}=\hat{X}^*$ into \eqref{ObjFofW2Problem}, it follows that
\begin{align}
\begin{split}
&\mathop {\min }\limits_{p_{Z, \hat{X}}} \ \alpha \mathbb{E} {\| {\hat{X} - G_1(Z)} \|^2} + (1-\alpha) W^{2}_{2}({p_{X'}},{p_{\hat{X}}})\\
=&\alpha (1-\alpha)\mathop {\min }\limits_{p_{X',Z}} \mathbb{E} {\left\|{G_1(Z) - X'} \right\|^2}\\
\ge&\alpha (1-\alpha) \mathbb{E} {\left\|{X_d - X} \right\|^2},\label{TakeXhatW2Problem}
\end{split}
\end{align}
where the last inequality is due to that $X_d=G_d(E_d(X))$ with $(E_d,G_d)$ being an MMSE encoder-decoder pair as defined in Theorem 1.
Now, we find an optimal encoder to \eqref{RelaxedW2Problem} given $\hat{X}^*$. From \eqref{TakeXhatW2Problem}, formulation \eqref{RelaxedW2Problem} can be written as
\begin{align}
\begin{split}
&\mathop {\min }\limits_{E\in \Omega} \ [\alpha \mathbb{E} {\left\|{G_1(E(X)) - X} \right\|^2} + \alpha (1-\alpha)\mathop {\min }\limits_{p_{X',E(X)}} \mathbb{E} {\left\|{G_1(E(X)) - X'} \right\|^2}]\\
\ge &\alpha \mathbb{E} {\left\|{X_d - X} \right\|^2} + \alpha (1-\alpha)\mathbb{E} {\left\|{X_d - X} \right\|^2}\\
= &\alpha (2-\alpha)\mathbb{E} {\left\|{X_d - X} \right\|^2}.\label{EncFunction}
\end{split}
\end{align}
Obviously, when $E=E_d$ and $p_{X',E(X)}=p_{X,Z_d}$, the equality in \eqref{EncFunction} holds. Hence, the MMSE encoder $E_d$ is optimal to \eqref{RelaxedW2Problem} for any $\alpha\in(0,1)$. 

Next, we consider a fixed encoder $E_d$ to show that 
$G^*_\alpha$ is an optimal decoder to \eqref{RelaxedW2Problem} for a fixed encoder $E_d$. 
When the encoder is fixed to $E_d$, from the definition of $(E_d,G_d)$, $X_d=G_d(E_d(X))$ and $G_1(Z)=\mathbb{E}[X|Z]$,
we have $G_1(E_d(X))=G_d(E_d(X))=X_d$. 
Then it follows from \eqref{OptimalXhat} that
\[\hat{X}^*=\alpha X_d + (1-\alpha)X',\]
with $X'$ satisfying $p_{X',Z_d}=p_{X,Z_d}$. 
Since the perfect perception decoder $G_p$ satisfies $p_{X_p,Z_d}=p_{X,Z_d}$ \cite{2021YanZy}, which means that $X'$ can be replaced by $X_p$ and it follows that
\[\hat{X}^*=\alpha X_d + (1-\alpha)X_p,\] 
is an optimal decoder to \eqref{RelaxedW2Problem} for fixed encoder $E_d$.
Therefore, with $Z_d=E_d(X)$, $X_d=G_d(E_d(X))$ and $X_p=G_p(E_d(X))$, $G^*_\alpha(Z_d)=\alpha G_d(Z_d) + (1-\alpha)G_p(Z_d)$ is an optimal decoder to \eqref{RelaxedW2Problem}. That is $(E_d,G^*_\alpha)$ is an optimal encoder-decoder pair to \eqref{RelaxedW2Problem}.

Now we prove that the encoder-decoder pair $(E_d,G^*_\alpha)$ is also optimal to \eqref{OriginProblem}.
Taking $G^*_\alpha$ back into \eqref{TakeXhatW2Problem} we have


\begin{align}
\begin{split}
&\alpha \mathbb{E} {\left\| {G^*_\alpha(Z_d) - G_d(Z_d)} \right\|^2} + (1-\alpha) W^{2}_{2}({p_X},{p_{G^*_\alpha(Z_d)}})\\
&=\alpha (1-\alpha) \mathop {\min }\limits_{p_{X,Z_d}} \mathbb{E} {\left\|{G_d(Z_d) - X} \right\|^2}\\
&=\alpha (1-\alpha) P_d, \label{DandPof4}
\end{split}
\end{align}
where $P_d=W^{2}_{2}(p_X,p_{G_d(Z_d)})$ as defined in Theorem 1.
It follows from (\ref{DandPof4}) that
\begin{align}
W^{2}_{2}({p_X},{p_{G^*_\alpha(Z_d)}})=\alpha ^2 P_d. \label{squaredW2}
\end{align}
Thus, with $\alpha=\sqrt{P/P_d}$ for any $0\!<\!P\!<\!P_d$, and for a fixed encoder $E_d$,
the optimal decoder $G^*_\alpha$ of \eqref{RelaxedW2Problem} satisfies 
\begin{equation}
W^2_2(p_X,p_{G^*_\alpha(Z_d)})=P. \label{squaredW2Gp}
\end{equation}
Consider $0<P<P_d$ in \eqref{OriginProblem}, and let $(E^{\bullet},G^{\bullet})$ be any optimal encoder-decoder pair to \eqref{OriginProblem}, then, with \eqref{squaredW2Gp}, for $\alpha=\sqrt{P/P_d}$ we have 
\begin{align}
\mathbb{E} {\left\| {X - G^{\bullet}(E^{\bullet}(X))} \right\|^2}&\le\mathbb{E} {\left\| {X - G^*_\alpha(E_d(X))} \right\|^2},\label{Prof2Dist}\\
W^2_2(p_X,p_{G^{\bullet}(E^{\bullet}(X))})&\le P= W^2_2(p_X,p_{G^*_\alpha(E_d(X))}). \label{Prof2eq3and41}
\end{align}
Summing up \eqref{Prof2Dist} and \eqref{Prof2eq3and41} yields
\begin{align}
\begin{split}
&\alpha\mathbb{E} {\left\| {X - G^{\bullet}(E^{\bullet}(X))} \right\|^2}+(1-\alpha)W^2_2(p_X,p_{G^{\bullet}(E^{\bullet}(X))})\\\le&\alpha\mathbb{E} {\left\| {X - G^*_\alpha(E_d(X))} \right\|^2}+(1-\alpha)W^2_2(p_X,p_{G^*_\alpha(E_d(X))}). \label{Prof2eq3and42}
\end{split}
\end{align}
Since $(E_d,G^*_\alpha)$ is an optimal encoder-decoder pair to \eqref{RelaxedW2Problem}, the equality in \eqref{Prof2eq3and42} holds. Hence, the equalities in \eqref{Prof2Dist} and \eqref{Prof2eq3and41} holds for any $(E^{\bullet},G^{\bullet})$, and $(E_d,G^*_\alpha)$ is an optimal solution to \eqref{OriginProblem}. 
This completes the proof of Theorem 1.

\section{Proof of Theorem 2.}

Denote $Y\!:=\!(X,X_d)$, $\hat{Y}\!:=\!(\hat{X},X_d)$ and $Y_d\!:=\!(X_d,X_d)$, with which \eqref{normOFun} can be rewritten as
\begin{align}
\mathop {\min }\limits_{{p_{\hat Y,Y_d}}} W_1(p_{\hat Y},p_{Y})+ \lambda \mathbb{E}\|\hat{Y}-Y_d \|.\label{YOFun}
\end{align}
Then, to justify Theorem 2 \textit{i)}, it is enough to justify that, 
when $0 \le \lambda<1$, the optimal solution of \eqref{YOFun} satisfies $p_{\hat Y}=p_{Y}$.
Let $p^*_{\hat{Y},Y_d}$ be an optimal solution to \eqref{YOFun}, 
of which the  optimal decoding output is $\hat{Y}^*$ and denote $\hat{Y}^*\!:=\!(\hat{X}^*,X_d)$.
Then $p^*_{\hat{Y},Y_d}$ is also optimal to
\begin{align}
\mathop {\min }\limits_{{p_{\hat Y,Y_d}}} \mathbb{E}\|\hat{Y}-Y_d \|,
~~~~{\rm{s.t.}} ~~~~{p_{\hat{Y}}}={p_{\hat{Y}^*}},\label{Y*Yd}
\end{align}
which implies $\mathbb{E}\|\hat{Y}^*-Y_d \|=W_1(p_{\hat{Y}^*}, p_{Y_d})$. Thus, the minimal objective value of \eqref{YOFun} is
\begin{align}
\begin{split}
L^* =& W_1(p_{\hat Y ^*},p_{Y})+ \lambda W_1(p_{\hat Y ^*},p_{Y_d})\\
\mathop {\rm{ \ge }}\limits^{(a)}& (1-\lambda)W_1(p_{\hat Y ^*},p_{Y})+\lambda W_1(p_{Y_d},p_{Y})\\
\mathop {\rm{ \ge }}\limits^{(b)}& \lambda W_1(p_{Y_d},p_{Y}).\label{optValue}
\end{split}
\end{align}
In (a) we use the property of Wasserstein distance and (b) is because of the non-negativity of $(1-\lambda)W_1(p_{\hat Y ^*},p_{Y})$. When $p_{\hat{Y}}=p_Y$, we have $W_1(p_{\hat Y ^*},p_{Y})=0$ and $L^*=\lambda W_1(p_{Y_d},p_{Y})$. The equality of (b) holds if and only if $W_1(p_{\hat Y ^*},p_{Y})=0$, which implies that the minimal objective is attained only when $p_{\hat{Y}^*}=p_Y$. Thus, $p_{\hat{Y}^*}=p_Y$ holds for any optimal solution $p^*_{\hat{Y},Y_d}$ of \eqref{YOFun}.
This leads to the result that, for any $0 \le \lambda<1$, the optimal solution of \eqref{normOFun} satisfies $p_{\hat X,X_d}=p_{X,X_d}$.

Next, we further justify that the optimal condition $p_{\hat X,X_d}=p_{X,X_d}$ is equivalent to $p_{\hat X,Z_d}=p_{X,Z_d}$. 
To this end, we show that
for an optimal MMSE decoder $G_d$, the mapping from $Z_d$ to $X_d$ is bijective. 
Specifically, given a $z_d$, its corresponding MMSE decoding $x_d$ is  $x_d=G_d(z_d)=\mathbb{E}[X|z_d]$. 
Besides, for any $x_d$ in $X_d$, there is a unique $z_d$ in $Z_d$ satisfying $x_d=G_d(z_d)$, otherwise $Z_d$ can be further compressed into a lower bit-rate without increasing the MSE distortion. 
This concludes the proof of Theorem 2 \textit{i)}.

Now, we prove that when $\lambda>1$, the optimal solution to \eqref{normOFun} is $\hat{X}=X_d$. Similarly, we rewrite \eqref{normOFun} as \eqref{YOFun}. It is enough to prove that when $\lambda > 1$, the optimal solution to \eqref{YOFun} is $\hat{Y}=Y_d$. For the optimal solution $p^*_{\hat{Y},Y_d}$, let the distribution of output be $p_{\hat{Y}^*}$. Since $\mathbb{E}\|\hat{Y}^*-Y_d \|=W_1(p_{\hat{Y}^*}, p_{Y_d})$, the minimal objective value of \eqref{YOFun} is
\begin{align}
\begin{split}
L^*=&W_1(p_{\hat Y},p_{Y})+ \lambda \mathbb{E}\|\hat{Y}-Y_d \|\\
=& W_1(p_{\hat Y ^*},p_{Y})+ \lambda W_1(p_{\hat Y ^*},p_{Y_d})\\
\mathop {\rm{ \ge }}& W_1(p_{Y_d},p_{Y})+(\lambda-1) W_1(p_{\hat{Y}^*},p_{Y_d})\\
\mathop {\rm{ \ge }}& W_1(p_{Y_d},p_{Y}).\label{optValue2}
\end{split}
\end{align}
Obviously, \eqref{normOFun} reaches the infimum $W_1(p_{Y_d},p_{Y})$ if and only if $W_1(p_{\hat{Y}^*},p_{Y_d})=0$, which result in $p_{\hat{Y}^*}=p_{Y_d}$. Hence, the objective function of \eqref{normOFun} becomes
\begin{align}
W_1(p_{Y_d},p_{Y})+ \lambda \mathbb{E}\|\hat{Y}-Y_d \|.
\end{align}
Then, we have the result that $\hat{Y}=Y_d$ is the only solution to \eqref{normOFun} when $\lambda>1$. This concludes the proof of Theorem 2 \textit{ii)}.

\section{Applying Framework B to Post-process the BPG Codec}

\begin{figure*}[!t]
	\vskip 0.2in
	\begin{center}
	\subfigure[Ground-truth (depth image)]{
			\begin{minipage}[t]{0.38\columnwidth}
				\centering
				\includegraphics[width=\columnwidth]{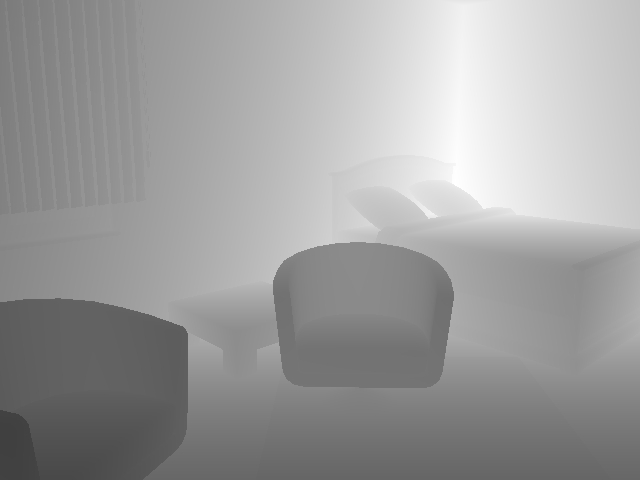}
			\end{minipage}%
		}%
		
	\subfigure[Ground-truth (point cloud)]{
			\begin{minipage}[t]{0.38\columnwidth}
				\centering
				\includegraphics[width=\columnwidth]{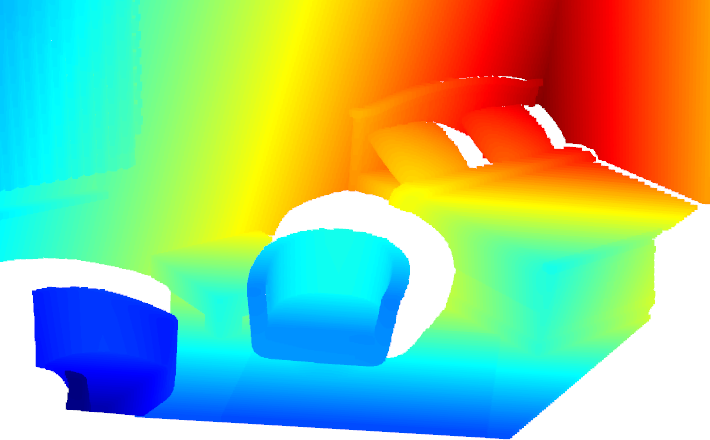}
			\end{minipage}%
			\label{suncgd}
		}~~
	\subfigure[BPG]{
			\begin{minipage}[t]{0.38\columnwidth}
				\centering
				\includegraphics[width=\columnwidth]{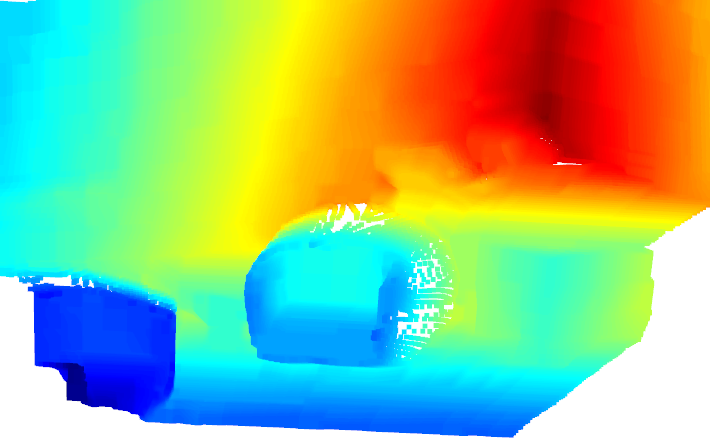}
			\end{minipage}%
		}%
		
	\subfigure[$X_\alpha$ with $\alpha=0$ (equivalent to $G_p$)]{
			\begin{minipage}[t]{0.38\columnwidth}
				\centering
				\includegraphics[width=\columnwidth]{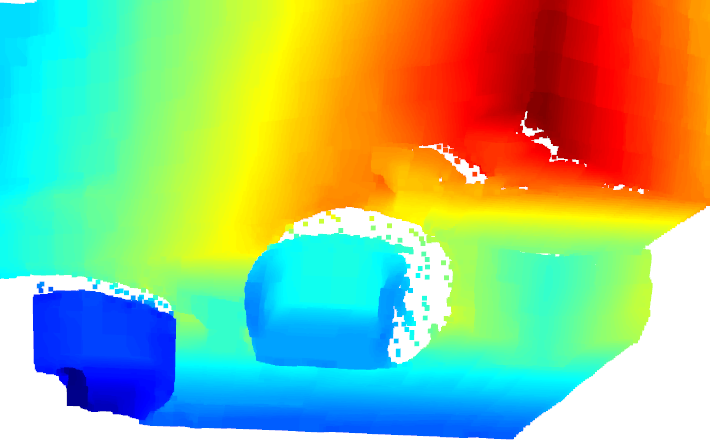}
			\end{minipage}%
			\label{suncgd}
		}~~~%
	\subfigure[$X_\alpha$ with $\alpha=0.25$]{
			\begin{minipage}[t]{0.38\columnwidth}
				\centering
				\includegraphics[width=\columnwidth]{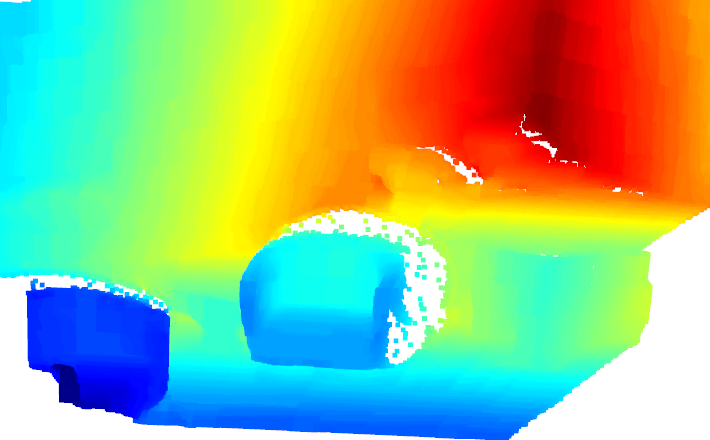}
			\end{minipage}%
		}%
		
	\subfigure[$X_\alpha$ with $\alpha=0.5$]{
			\begin{minipage}[t]{0.38\columnwidth}
				\centering
				\includegraphics[width=\columnwidth]{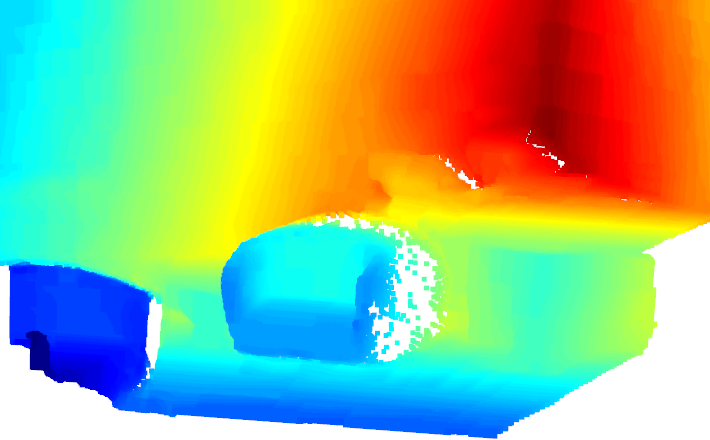}
			\end{minipage}%
		}~~~%
	\subfigure[$X_\alpha$ with $\alpha=0.75$]{
			\begin{minipage}[t]{0.38\columnwidth}
				\centering
				\includegraphics[width=\columnwidth]{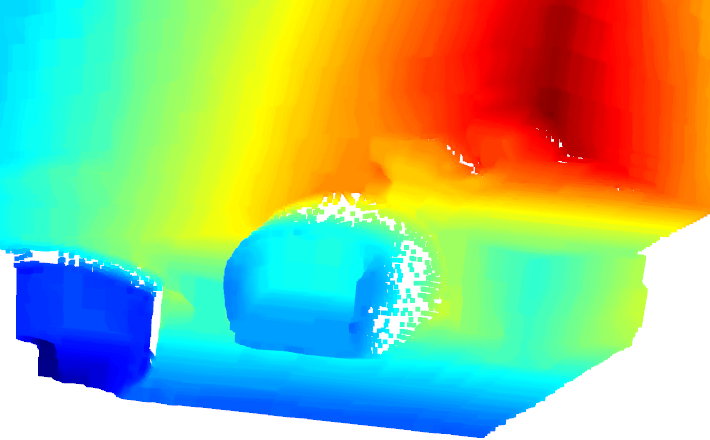}
			\end{minipage}%
		}%
		
	\caption{An illustration of applying Framework B to Post-process the BPG Codec on the SUNCG dataset. (a) Ground-truth depth image. (b) Point cloud corresponding to (a). (c) Point cloud reconstructed from the output of BPG. (d)-(g) Interpolation results. For visual clarity, the point cloud results are provided instead of depth image results.}
		\label{suncgbpg}
	\end{center}
	\vskip -0.2in
\end{figure*}

Section 5.2 has provided results on depth images by training a 
post-processing perceptual decoding network (i.e., the proposed 
Framework B as shown in Figure~\ref{postprocess}) to learn the 
distribution of the source conditioned on the MMSE decoding. 
From Theorem 2, Framework B is optimal if ($E_d$,$G_d$) is an 
MMSE encoder-decoder pair, which is a strict condition in practice. 
However, this framework applies to any existing compression system
that optimized only in terms of distortion (may be not optimized in terms of MSE). In fact, there exist many traditional lossy 
compression systems that do not consider perceptual decoding but only optimized to minimize certain distortion measures.

Here, we apply Framework B to post-process the BPG codec. 
We provide experimental results on depth images, in which
the encoder-decoder pair $(E_d,G_d)$ in Framework B is replaced by
BPG codec. BPG (QP=40) is used to replace $(E_d, G_d)$ in Figure~\ref{postprocess}, and we train a post-processing network $G_p$ to 
improve the perceptual quality of the BPG output. The parameters 
are set the same as in Section 5.2. 

Figure~\ref{suncgbpg} shows the results on a typical sample from the SUNCG dataset. 
For visual clarity, the point cloud results are provided instead of depth image results.
Although BPG is not optimal in MSE, our framework still shows high effectiveness by 
post perceptual decoding. The outputs of $G_p$ is much closer to the ground-truth than 
that of BPG.


\section{Result on RGB Images}

\begin{figure*}[!t]
	\begin{center}
		\subfigure[Ground-truth]{
			\begin{minipage}[t]{0.2775\columnwidth}
				\centering
				\includegraphics[width=\columnwidth]{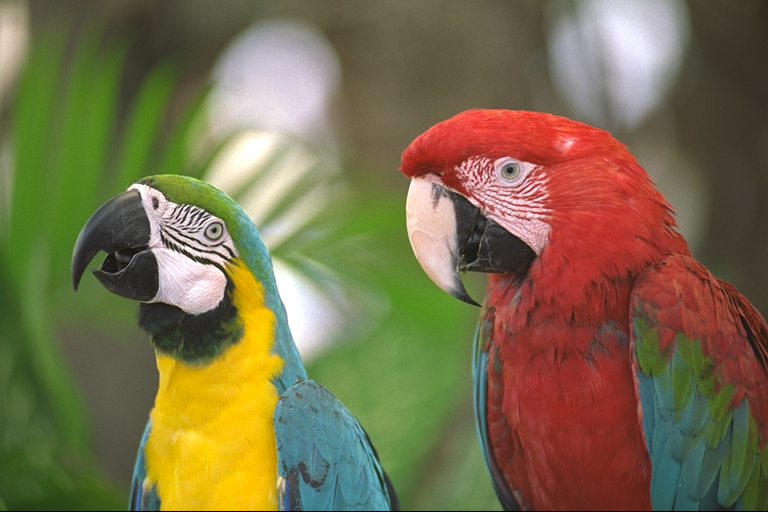}
			\end{minipage}%
			\hspace{1mm}%
			\begin{minipage}[t]{0.185\columnwidth}
				\centering
				\includegraphics[width=\columnwidth]{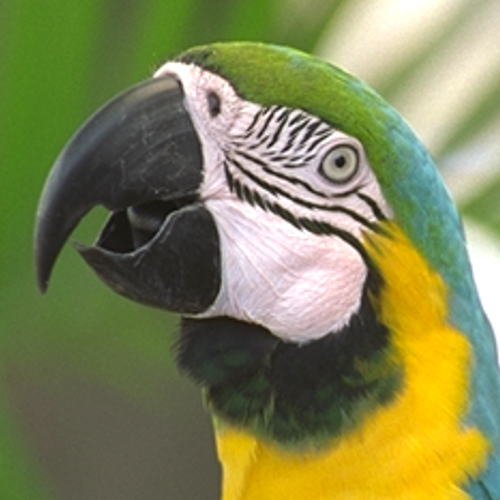}
			\end{minipage}%
		}%
		\subfigure[$G_d$ (MMSE)]{
			\begin{minipage}[t]{0.185\columnwidth}
				\centering
				\includegraphics[width=\columnwidth]{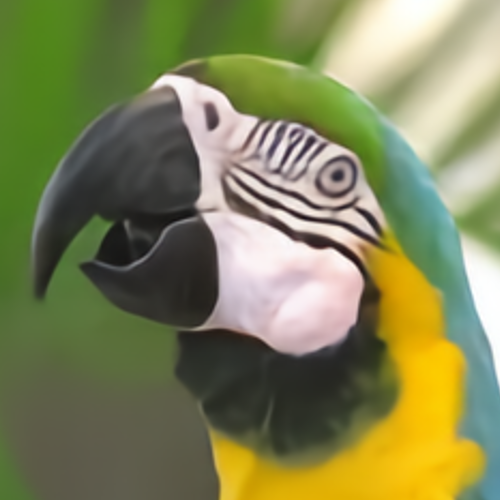}
			\end{minipage}%
		}%
		\subfigure[HiFiC]{
			\begin{minipage}[t]{0.185\columnwidth}
				\centering
				\includegraphics[width=\columnwidth]{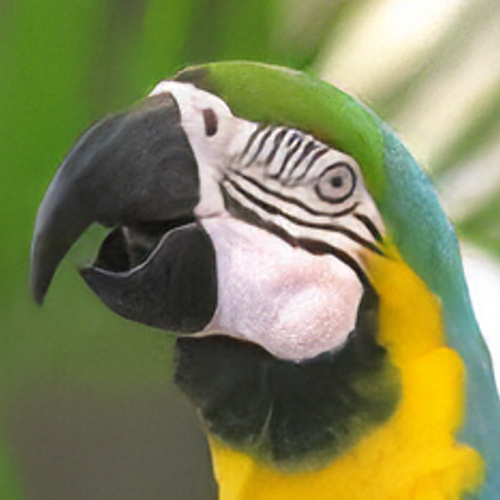}
			\end{minipage}%
		}%
		
		\subfigure[Framework A ($\alpha=0$ and $\alpha=0.5$)]{
			\begin{minipage}[t]{0.185\columnwidth}
				\centering
				\includegraphics[width=\columnwidth]{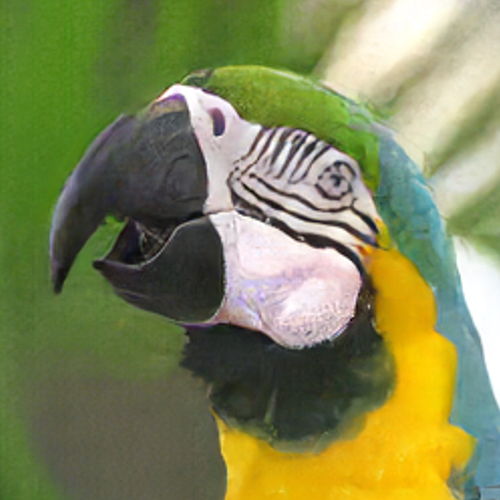}
			\end{minipage}%
			\hspace{1mm}%
			\begin{minipage}[t]{0.185\columnwidth}
				\centering
				\includegraphics[width=\columnwidth]{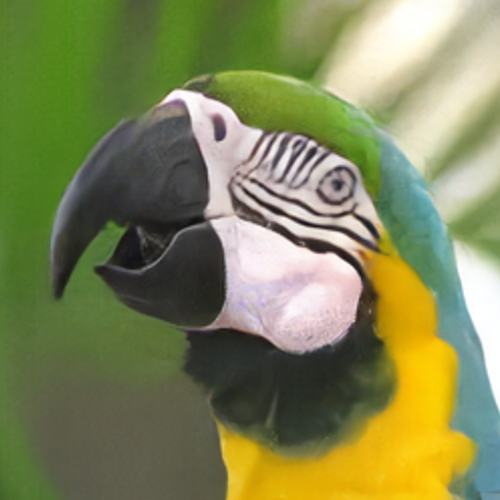}
			\end{minipage}%
		}\hspace{4mm}%
		\subfigure[Framework B ($\alpha=0$ and $\alpha=0.5$)]{
			\begin{minipage}[t]{0.185\columnwidth}
				\centering
				\includegraphics[width=\columnwidth]{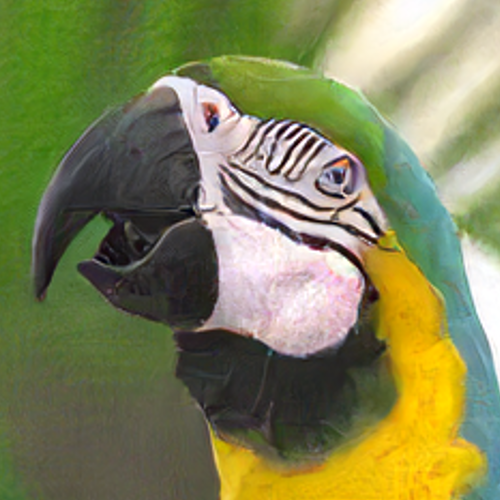}
			\end{minipage}%
			\hspace{1mm}%
			\begin{minipage}[t]{0.185\columnwidth}
				\centering
				\includegraphics[width=\columnwidth]{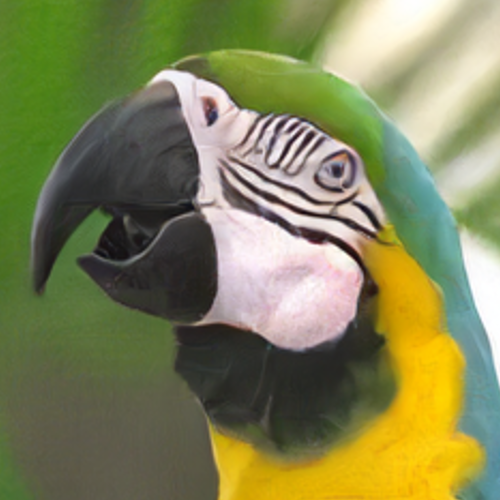}
			\end{minipage}%
		}%
		\caption{Example-1: Decoding outputs of the compared methods on a sample form the KODAK dataset.}
		\label{KODAK-1}
	\end{center}
\end{figure*}

\begin{figure*}[!t]
	\begin{center}
		\subfigure[Ground-truth]{
			\begin{minipage}[t]{0.2775\columnwidth}
				\centering
				\includegraphics[width=\columnwidth]{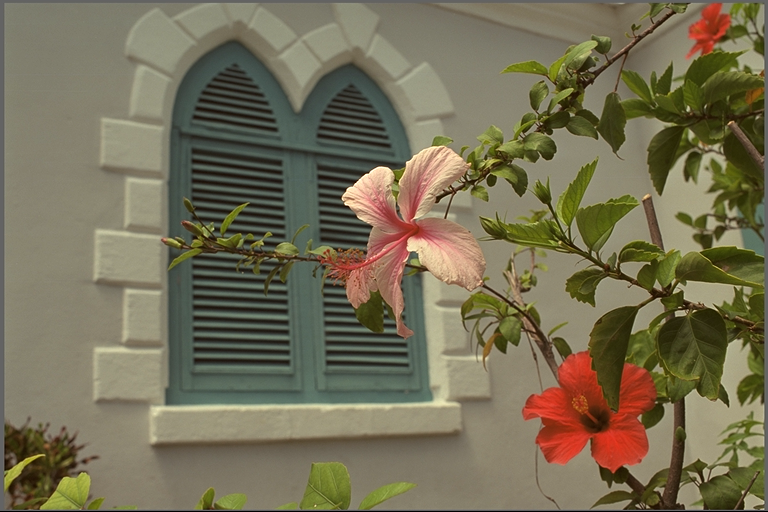}
			\end{minipage}%
			\hspace{1mm}%
			\begin{minipage}[t]{0.178\columnwidth}
				\centering
				\includegraphics[width=\columnwidth]{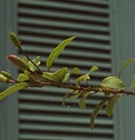}
			\end{minipage}%
		}%
		\subfigure[$G_d$ (MMSE)]{
			\begin{minipage}[t]{0.178\columnwidth}
				\centering
				\includegraphics[width=\columnwidth]{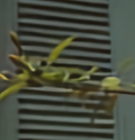}
			\end{minipage}%
		}%
		\subfigure[HiFiC]{
			\begin{minipage}[t]{0.178\columnwidth}
				\centering
				\includegraphics[width=\columnwidth]{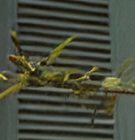}
			\end{minipage}%
		}%
		
		\subfigure[Framework A ($\alpha=0$ and $\alpha=0.5$)]{
			\begin{minipage}[t]{0.178\columnwidth}
				\centering
				\includegraphics[width=\columnwidth]{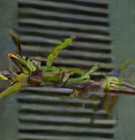}
			\end{minipage}%
			\hspace{1mm}%
			\begin{minipage}[t]{0.178\columnwidth}
				\centering
				\includegraphics[width=\columnwidth]{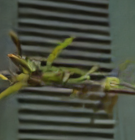}
			\end{minipage}%
		}\hspace{4mm}%
		\subfigure[Framework B ($\alpha=0$ and $\alpha=0.5$)]{
			\begin{minipage}[t]{0.178\columnwidth}
				\centering
				\includegraphics[width=\columnwidth]{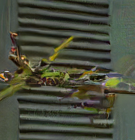}
			\end{minipage}%
			\hspace{1mm}%
			\begin{minipage}[t]{0.178\columnwidth}
				\centering
				\includegraphics[width=\columnwidth]{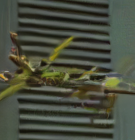}
			\end{minipage}%
		}%
		\caption{Example-2: Decoding outputs of the compared methods on a sample form the KODAK dataset.}
		\label{KODAK-2}
	\end{center}
\end{figure*}

\begin{figure*}[!t]
	\begin{center}
		\subfigure[Ground-truth]{
			\begin{minipage}[t]{0.28\columnwidth}
				\centering
				\includegraphics[width=\columnwidth]{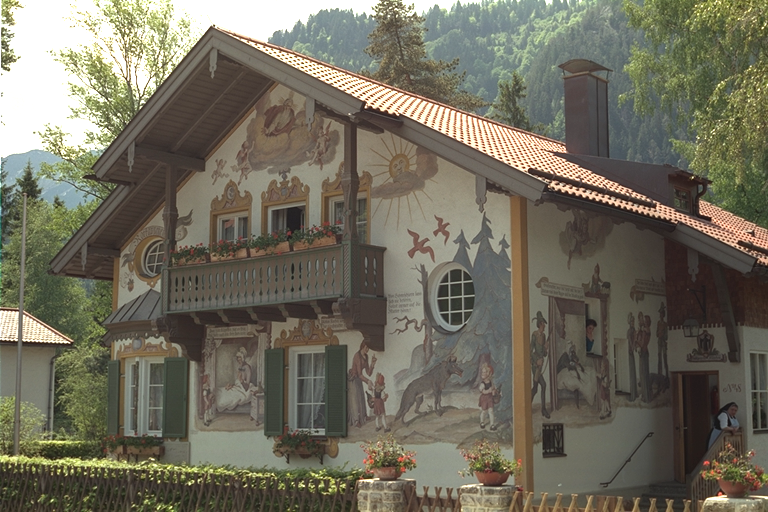}
			\end{minipage}%
			\hspace{1mm}%
			\begin{minipage}[t]{0.188\columnwidth}
				\centering
				\includegraphics[width=\columnwidth]{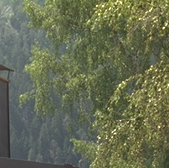}
			\end{minipage}%
		}%
		\subfigure[$G_d$ (MMSE)]{
			\begin{minipage}[t]{0.188\columnwidth}
				\centering
				\includegraphics[width=\columnwidth]{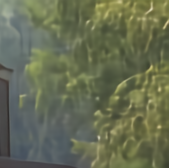}
			\end{minipage}%
		}%
		\subfigure[HiFiC]{
			\begin{minipage}[t]{0.188\columnwidth}
				\centering
				\includegraphics[width=\columnwidth]{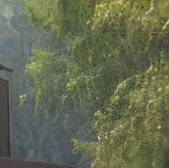}
			\end{minipage}%
		}%
		
		\subfigure[Framework A ($\alpha=0$ and $\alpha=0.5$)]{
			\begin{minipage}[t]{0.188\columnwidth}
				\centering
				\includegraphics[width=\columnwidth]{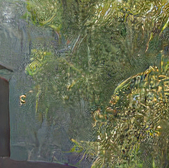}
			\end{minipage}%
			\hspace{1mm}%
			\begin{minipage}[t]{0.188\columnwidth}
				\centering
				\includegraphics[width=\columnwidth]{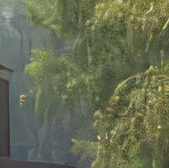}
			\end{minipage}%
		}\hspace{3mm}%
		\subfigure[Framework B ($\alpha=0$ and $\alpha=0.5$)]{
			\begin{minipage}[t]{0.188\columnwidth}
				\centering
				\includegraphics[width=\columnwidth]{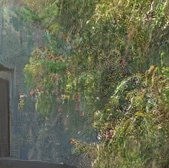}
			\end{minipage}%
			\hspace{1mm}%
			\begin{minipage}[t]{0.188\columnwidth}
				\centering
				\includegraphics[width=\columnwidth]{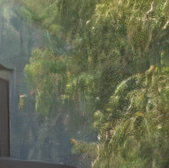}
			\end{minipage}%
		}%
		
		\caption{Example-3: Decoding outputs of the compared methods on a sample form the KODAK dataset.}
		\label{KODAK-3}
	\end{center}
\end{figure*}

Here, we provide more samples for qualitative comparison on RGB images.
The image samples are from in KODAK dataset, as shown in Figure~\ref{KODAK-1}-Figure~\ref{KODAK-3}.
The state-of-the-art method HiFiC is compared.
Compared with the MMSE decoder $G_d$, the outputs of both 
Framework A and Framework B contain more details and the edges are sharper. 


Besides, we also study the effect of $\lambda$ in 
\eqref{normOFun} on the practical performance of 
the proposed Framework A. Theoretically, 
by Theorem 2, for any $\lambda < 1$, optimizing \eqref{normOFun} 
leads to perfect perception decoding. However, as explained in 
Remark 5, the discriminator in WGAN-gp is not strictly 1-Lipschitz 
and the capacity of the used network is not infinite.
Hence, in practice the optimal solution cannot be achieved, 
especially when data distribution is complex, e.g., RGB images. 
We test different values of $\lambda$ when training the proposed 
Framework A. The quantitative results on the KODAK dataset are 
provided in Table~\ref{lambdatable},
including the PSNR and three perception indices. 
It can be seen that, as the value of $\lambda$ decreases,
the PSNR decreases but each perception index improves.

\begin{figure*}[!t]
	\vskip 0.2in
	\begin{center}
		\subfigure[Ground-truth]{
			\begin{minipage}[t]{0.24\columnwidth}
				\centering
				\includegraphics[width=\columnwidth]{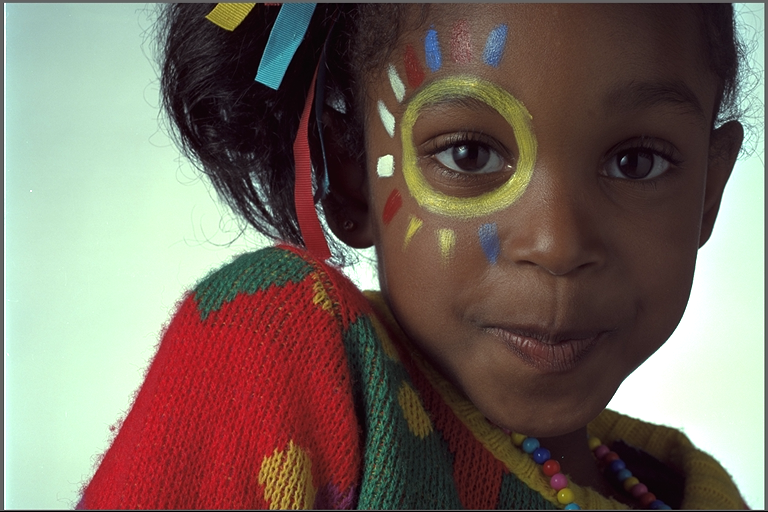}
			\end{minipage}%
		}%
		\subfigure[MMSE]{
			\begin{minipage}[t]{0.24\columnwidth}
				\centering
				\includegraphics[width=\columnwidth]{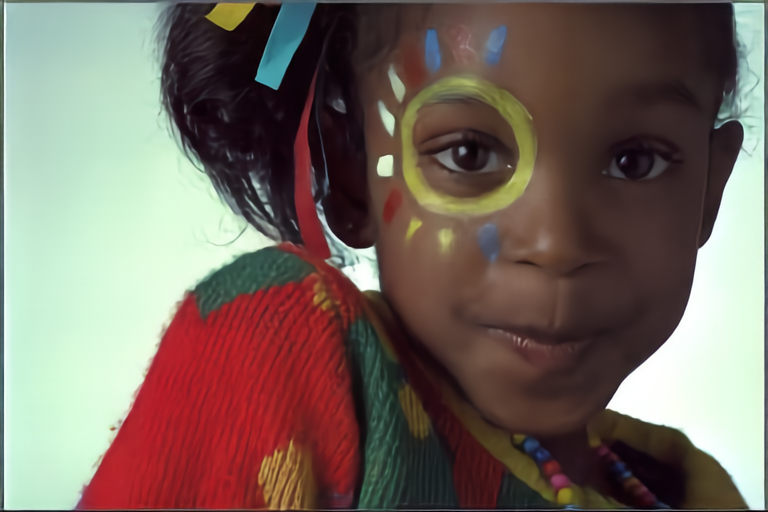}
			\end{minipage}%
		}%
		\subfigure[$\lambda=0.4$]{
			\begin{minipage}[t]{0.24\columnwidth}
				\centering
				\includegraphics[width=\columnwidth]{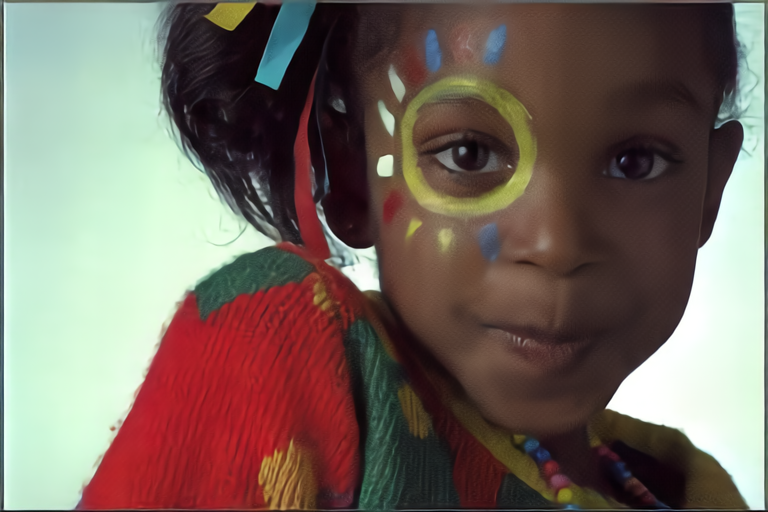}
			\end{minipage}%
		}%
		\subfigure[$\lambda$=0.2]{
			\begin{minipage}[t]{0.24\columnwidth}
				\centering
				\includegraphics[width=\columnwidth]{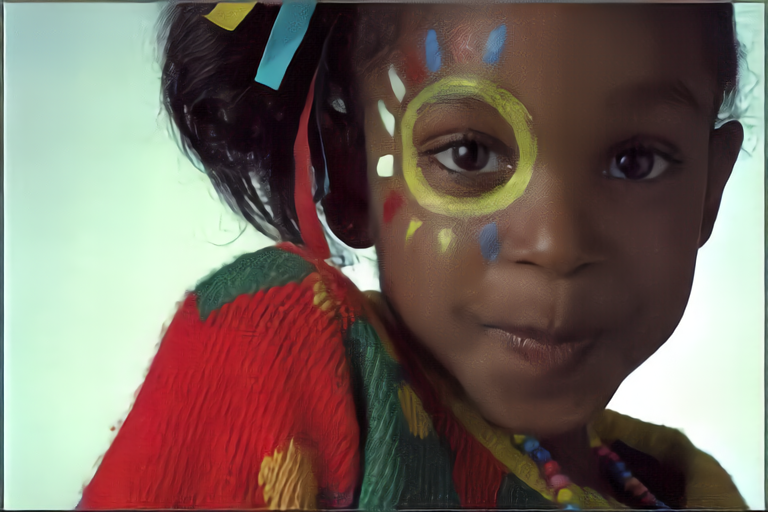}
			\end{minipage}%
		}%
		
		\subfigure[$\lambda$=0.1]{
			\begin{minipage}[t]{0.24\columnwidth}
				\centering
				\includegraphics[width=\columnwidth]{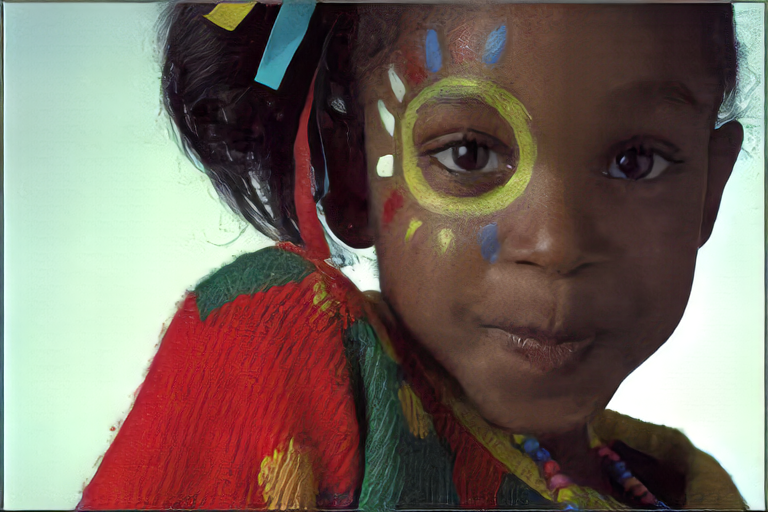}
			\end{minipage}%
		}%
		\subfigure[$\lambda$=0.05]{
			\begin{minipage}[t]{0.24\columnwidth}
				\centering
				\includegraphics[width=\columnwidth]{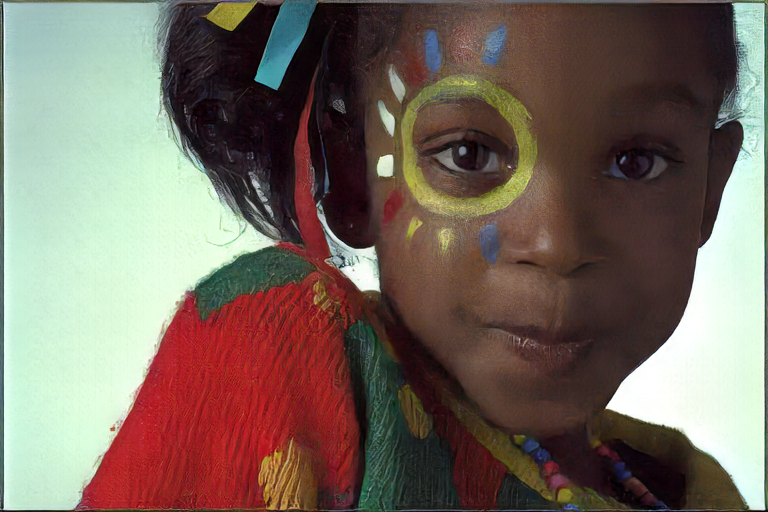}
			\end{minipage}%
		}%
		\subfigure[$\lambda$=0.01]{
			\begin{minipage}[t]{0.24\columnwidth}
				\centering
				\includegraphics[width=\columnwidth]{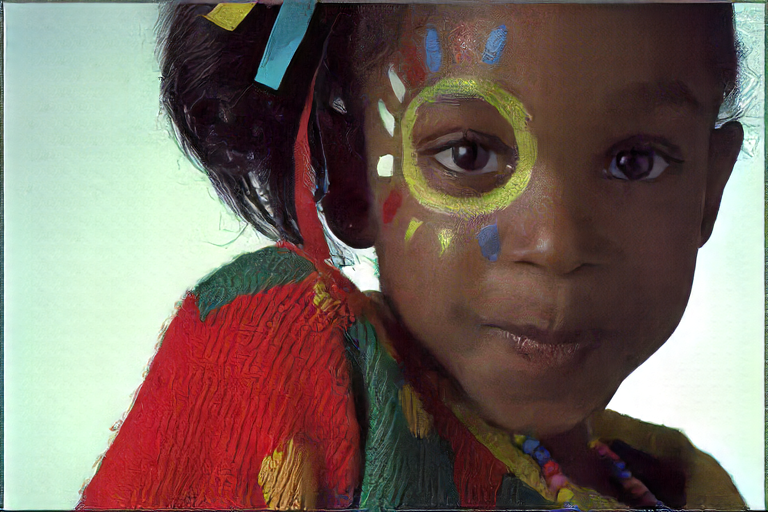}
			\end{minipage}%
		}%
		\subfigure[$\lambda$=0.005]{
			\begin{minipage}[t]{0.24\columnwidth}
				\centering
				\includegraphics[width=\columnwidth]{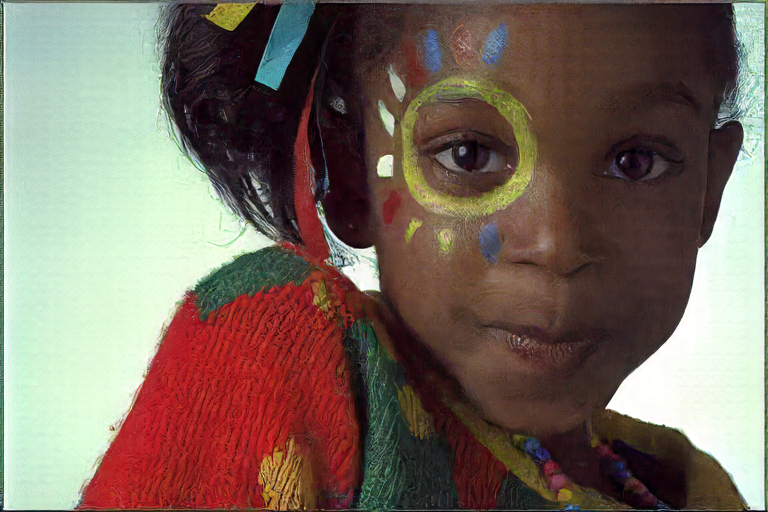}
			\end{minipage}%
		}%
		
		\caption{Decoding outputs of the proposed Framework A with different $\lambda$ on a typical sample from the KODAK dataset.}
		\label{lambdasample}
	\end{center}
	\vskip -0.2in
\end{figure*}

Figure~\ref{lambdasample} presents the decoding outputs of 
the proposed Framework A with different $\lambda$ on a typical sample.
As shown in Figure~\ref{lambdasample}, when $\lambda \ge 0.2$, 
the visual quality of the decoding outputs is similar to that of the MMSE decoding. 
When $\lambda \le 0.1$, the visual quality improves significantly  
as more conspicuous details can be observed, but artifacts also increase.

Since the tuning of adversarial training is much involved,
we believe that the perceptual quality of the proposed frameworks
can be further improved by more intensive tuning of the hyper-parameters 
and network architecture.

\begin{table}[t]
\caption{RGB image decoding results by the proposed Framework A for different values of $\lambda$ in \eqref{normOFun} (on the KODAK dataset).}
\label{sample-table}
\vskip 0.15in
\begin{center}
\begin{small}
\begin{sc}
\begin{tabular}{lcccr}
\toprule
$\lambda$ & PSNR & PI & Ma & NIQE \\
\midrule
MMSE & 38.36& 4.54 & 6.36 & 5.45\\
0.4    & 38.27 & 3.93	& 6.82 & 4.68 \\
0.2    & 37.73 & 3.09	& 7.68 &	3.78\\
0.1    & 37.22 & 2.45	& 8.26	& 3.16\\
0.05   & 37.18 & 2.12 & 8.60 & 2.84\\
0.01   & 36.89 & 2.05	& 8.74	& 2.84\\
0.005   & 36.96 & 1.98 & 8.75 &	2.71\\
\bottomrule
\label{lambdatable}
\end{tabular}
\end{sc}
\end{small}
\end{center}
\vskip -0.1in
\end{table}

\end{document}